\documentclass[letterpaper]{article} 
\usepackage{aaai25}  
\usepackage{times}  
\usepackage{helvet}  
\usepackage{courier}  
\usepackage[hyphens]{url}  
\usepackage{graphicx} 
\usepackage{natbib}  
\usepackage{caption} 
\usepackage[algo2e, ruled,vlined]{algorithm2e}
\usepackage[utf8]{inputenc} 
\usepackage[T1]{fontenc}    
\usepackage{url}            
\usepackage{booktabs}       
\usepackage{amsfonts}       
\usepackage{nicefrac}       
\usepackage{microtype}      
\usepackage{xcolor}         

\usepackage{booktabs}       
\usepackage{amsfonts}       
\usepackage{nicefrac}       
\usepackage{microtype}      
\usepackage{fancyhdr}       
\usepackage{graphicx}       

\usepackage[para]{threeparttable}

\usepackage{natbib}

\usepackage{microtype}
\usepackage{graphicx,fancybox}
\usepackage{subfigure}
\usepackage{booktabs} 
\usepackage{algorithm,algorithmic}
\usepackage{amssymb, multirow, paralist, color,amsmath,amsthm}

\usepackage{enumitem}
\setlist[itemize]{align=parleft,left=1em}
\usepackage{textcase}
\usepackage[T1]{fontenc}
\usepackage{booktabs} 
\usepackage{makecell}
\usepackage{comment}
\def \S {\mathbf{S}}

\def \1 {\mathbf{1}}

\def \S {\mathcal{S}}

\newtheorem{definition}{Definition}

\usepackage{newfloat}
\usepackage{listings}

\DeclareCaptionStyle{ruled}{labelfont=normalfont,labelsep=colon,strut=off} 
\floatstyle{ruled}
\newfloat{listing}{tb}{lst}{}
\floatname{listing}{Listing}
\title{GeoPro-Net: Learning Interpretable Spatiotemporal Prediction Models through Statistically-Guided Geo-Prototyping}
\author {
    Bang An\textsuperscript{\rm 2}\equalcontrib ,
    Xun Zhou\textsuperscript{\rm 1}\equalcontrib\thanks{Corresponding authors.},
    Zirui Zhou\textsuperscript{\rm 1},
    Ronilo Ragodos\textsuperscript{\rm 3},
    Zenglin Xu\textsuperscript{\rm 4},
    Jun Luo\textsuperscript{\rm 5}
}
\affiliations {
    \textsuperscript{\rm 1}School of Computer Science and Technology, Harbin Institute of Technology, Shenzhen\\
    \textsuperscript{\rm 2}Scheller College of Business, Georgia Institute of Technology\\
    \textsuperscript{\rm 3}Tippie College of Business, University of Iowa\\
    \textsuperscript{\rm 4}Artificial Intelligence Innovation and Incubation Institute, Fudan University\\
    \textsuperscript{\rm 5}Logistics and Supply Chain MultiTech R\&D Centre\\

    bang.an@scheller.gatech.edu,
    zhouxun2023@hit.edu.cn, 23s151098@stu.hit.edu.cn, ronilo-ragodos@uiowa.edu, zenglinxu@fudan.edu.cn, jluo@lscm.hk
}

\begin{document}

\maketitle

\begin{abstract}
The problem of forecasting spatiotemporal events such as crimes and accidents is crucial to public safety and city management. Besides accuracy, interpretability is also a key requirement for spatiotemporal forecasting models to justify the decisions. Interpretation of the spatiotemporal forecasting mechanism is, however, challenging due to the complexity of multi-source spatiotemporal features, the non-intuitive nature of spatiotemporal patterns for non-expert users, and the presence of spatial heterogeneity in the data.
Currently, no existing deep learning model intrinsically interprets the complex predictive process learned from multi-source spatiotemporal features. To bridge the gap, we propose GeoPro-Net, an intrinsically interpretable spatiotemporal model for spatiotemporal event forecasting problems. GeoPro-Net introduces a novel Geo-concept convolution operation, which employs statistical tests to extract predictive patterns in the input as ``Geo-concepts'', and condenses the ``Geo-concept-encoded'' input through interpretable channel fusion and geographic-based pooling. In addition, GeoPro-Net learns different sets of prototypes of concepts inherently, and projects them to real-world cases for interpretation. Comprehensive experiments and case studies on four real-world datasets demonstrate that GeoPro-Net provides better interpretability while still achieving competitive prediction performance compared with state-of-the-art baselines.
\end{abstract}

%

\section{Introduction}

The spatiotemporal event forecasting problem aims at predicting where and when certain features or events will occur in the geographic space and time. The applications of spatiotemporal event forecasting such as crime prediction and traffic accident forecasting hold significant importance in various domains, including public safety\cite{CPD} and traffic management\cite{acctool}. 
While it is important to make accurate predictions, understanding the predictive process of models in spatiotemporal event forecasting is, if not more, equally important to a wide range of urban stakeholders. This approach assists in justifying government expenditure on unforeseen risks, thereby contributing to more informed decision-making in urban contexts.

Interpreting spatiotemporal event prediction models is a challenging task. Firstly, spatiotemporal event prediction often involves multi-source input, such as traffic volume, precipitation, and point-of-interests, with complex dependencies and semantics. Interpreting their interplay over space and time is a non-trivial task. Secondly, unlike other problems such as image classification, where a picture can be naturally understood by humans, visualizing high-dimensional spatiotemporal features with post-hoc analysis on a map (e.g., saliency map) is not readily understandable by most individuals without domain knowledge. Thirdly, spatial heterogeneity \cite{10.1145/3161602} commonly exists in spatiotemporal event datasets, which implies that the rationale behind the event occurrences might vary significantly from place to place. It poses difficulty in interpreting the modeling process within a heterogeneous space. These challenges call for an \emph{intrinsically interpretable} model specifically designed for \emph{spatiotemporal event forecasting problems}. 

However, to the best of our knowledge, \textbf{there has not been any intrinsically interpretable deep learning model for the spatiotemporal event forecasting problem}. \emph{On the spatiotemporal event model side}, early machine learning methods \cite{DrukkerDavidM.2013MLaG, BrunsdonChris1996GWRA, JiangZhe2015LSDT} leveraged regression-based or tree-based models for interpretability but rely on manual feature augmentation, therefore, struggle to capture complex spatiotemporal dependencies in the data. Attention-based spatiotemporal event forecasting methods \cite{liu20120dynamic,guo2019attention} generate visualizations of scores indicating relevance between input and target locations but fall short of offering insights into the underlying reasons for events at the feature level. 
\emph{On the interpretable machine learning side} prototype-based explainers \cite{NEURIPS2019_adf7ee2d} are both intrinsically interpretable and naturally understandable by humans when used for image classification. However, visualizing multi-source spatiotemporal features in the same way makes it difficult to be understood. To strengthen the interpretability for abstract and complex features, concept-based models \cite{pmlr-v119-koh20a} have been introduced to extract high-level information from inputs, making it easier to understand. However, training data must be manually annotated with predefined but interpretable concepts. 

In this paper, we bridge the research gap 
by 
proposing the first intrinsically interpretable deep learning model for spatiotemporal event prediction, named \textbf{GeoPro-Net}. As a classifier, GeoPro-Net predicts if an event will occur at a given geo-location and time based on the spatiotemporal features observed around the location. 
To unravel the complexity of multi-source spatiotemporal features and transform them into comprehensible insights, we introduce a statistically guided retrieval process to identify multi-scale patterns such as local and global anomalies, thereby filtering noisy spatiotemporal information and alleviating the interpretation difficulty. Furthermore, we incorporate the idea of concepts \cite{pmlr-v119-koh20a} and structuralize the captured spatiotemporal event patterns into geographic-based concepts (Geo-concepts) via a novel multi-scale hierarchical pooling mechanism. This enables non-expert users to easily understand the extracted and detailed spatial information. Lastly, we learn a set of prototypes, which are representative and predictive combinations of Geo-concepts associated with the two output classes. These prototypes generate predictions through a linear layer and can be projected onto cases in the training set, making predictions interpretable with real-world examples. Comprehensive experiments and case studies on four real-world event datasets in Chicago and New York City demonstrate the faithful interpretability and better prediction accuracy of GeoPro-Net in comparison with other baselines. The code is available at https://github.com/BANG23333/GeoProNet
\underline{Our main contributions are summarized below}:
\begin{itemize}
    \setlength\itemsep{-0.0em}
    \item To the best of our knowledge, this is the first prototype-based interpretable framework to learn from spatiotemporal event datasets and explain its reasoning process intrinsically.
    \item We propose a novel spatial conceptualization process to extract Geo-concepts, which are statistically interpretable multi-scale input feature patterns, followed by condensing concepts through channel fusion and geographic-based multi-scale hierarchical pooling. 
    \item We propose to use a prototype learning framework to obtain a set of Geo-concepts associated with the occurrence or absence of events. By projecting these prototypes into training cases, we interpret predictions in the context of real-world scenarios.

\end{itemize}

\vspace{-2mm}

\section{Related Work}


\textbf{Interpretable Spatiotemporal Event Forecasting Models} traditionally rely on regression based methods (ex. Spatial Autoregressive Model\cite{BrunsdonChris1996GWRA} and Geographically Weighted Regression \cite{DrukkerDavidM.2013MLaG}) and tree-based methods (ex. Spatial Decision Tree \cite{JiangZhe2015LSDT}) on small-scale datasets with limited features. Those methods are inherently explainable via their learned coefficients or tree structures. However, their accuracy is largely limited when complicated spatiotemporal dependencies are involved. Deep learning approaches with enormous parameters \cite{6894591,yao2018deep,10.1145/3292500.3330884,AnBang2023SUER,MengLili2019ISAf,ZuoSimiao2021THP} has been proven effective with superior performance. However, most of the existing spatiotemporal event forecasting deep learning approaches are either black-box models or only partially interpretable. For example, xGAIL \cite{PanMenghai2020xEGA} was proposed to use GAIL-based methods \cite{10.5555/3157382.3157608} to draw insights into the decision-making process of taxi drivers. Furthermore, attention-based methods \cite{guo2019attention,liu20120dynamic,TangJiabin2023ESGN,DingYukai2020IsaL} are used to interpret the predictive process of forecasting urban events using spatial attention matrix and saliency maps. 

\textbf{Prototype-based Interpretable Models} \cite{NIPS2017_cb8da676,NEURIPS2019_adf7ee2d} have brought more attention in recent years. Most existing prototype-based interpretable models \cite{NEURIPS2019_adf7ee2d,NautaMeike2020NPTf}  are limited to image problems, where prototypes are projected as image tensors with RGB channels. Such prototypes are naturally human-understandable. For example, in a bird classification problem, people can easily distinguish different species parts by comparing birds' color, fur, paws, etc. However, such multi-dimensional prototype representations in urban incident problems are not directly interpretable from a human's perspective because a set of traffic speed, volume, and occupancy changing over time and space is not intuitive and self-explainable. 

\textbf{Concept-Based Interpretable Models} label their inputs with high-level concepts. Koh et al. \cite{koh2020concept} applied this approach by using bone spurs to predict arthritis in X-ray images. These concepts can either be predefined before training or learned during the training process. Essentially, concepts serve as extracted summaries of complex information. The systematic definition of necessary concepts is crucial, as having too few concepts may limit model performance, while an excessive number of concepts can compromise model interpretability.

\section{Preliminaries}

A $spatial$-$temporal$ $field$ $S \times T$ is a three-dimensional matrix, where $T = \{t_{1}, t_{2},...,t_{i}\}$ is a study period divided into equal length intervals (e.g., hours, days) and $S = \{s_{(0, 0)}, s_{(0, 1)},...,s_{(m, n)}\}$ is a $m \times n$ two-dimension spatial grid partitioned from the study area. Temporal features $F_T \in \mathbb{R}^{T \times f_{t}}$ (e.g., average temperature, day of week), spatial features $F_S \in \mathbb{R}^{m \times n \times f_{s}}$ (e.g., number of POIs), and spatio-temporal features $F_{ST} \in \mathbb{R}^{T \times m \times n \times f_{st}}$ (e.g., traffic volume) are mapped to $S\times T$ field. We denote the total number of features as $f = f_s+f_t + f_{st}$. The detailed list of features,  processing steps, and symbol table in this work can be found in Appendix A.

\textbf{Problem Definition: Given} the socio-environmental features $F_T$, $F_S$, $F_{ST}$ of a location $l_m$ and its neighboring locations (e.g., within the distance of $r$ grid cells) in time window $ t \in T $, our problem is to \textbf{Predict} whether an event will occur in the future one-step interval $t_{i+1}$ for location $l_m$. The \textbf{Objective} is to interpret the predictive process while minimizing prediction errors. 
In this spatiotemporal event forecasting problem, we work with a basic assumption commonly adopted by prior work that there exists spatial auto-correlation over space, meaning nearby locations tend to have more correlated values \cite{ToblerW.R.1970ACMS}.

\section{Methodology}
In this section, we present our GeoPro-Net model to intrinsically interpret the predictive process on the defined spatiotemporal event forecasting problems. Figure \ref{fig: framework} demonstrates the proposed model architecture.

\subsection{Statistically-guided Geo-Concept Encoding}
In cognitive science, the simplicity principle states that the mind tends to seek the simplest available interpretation of observations over complex ones \cite{JacobFeldman2016Tspi}.
It becomes crucial to extract the most representative information from extensive feature values and consolidate them into basic concepts to facilitate the interpretation of the modeling process. To address such challenges, we design a novel Statistically-guided Spatial Concept Encoding layer (SSCE) as demonstrated in the black dashed box of Figure \ref{fig: framework}.

\begin{figure}[ht] \hspace*{0cm}
 \centering
 \includegraphics[width = 0.47\textwidth]{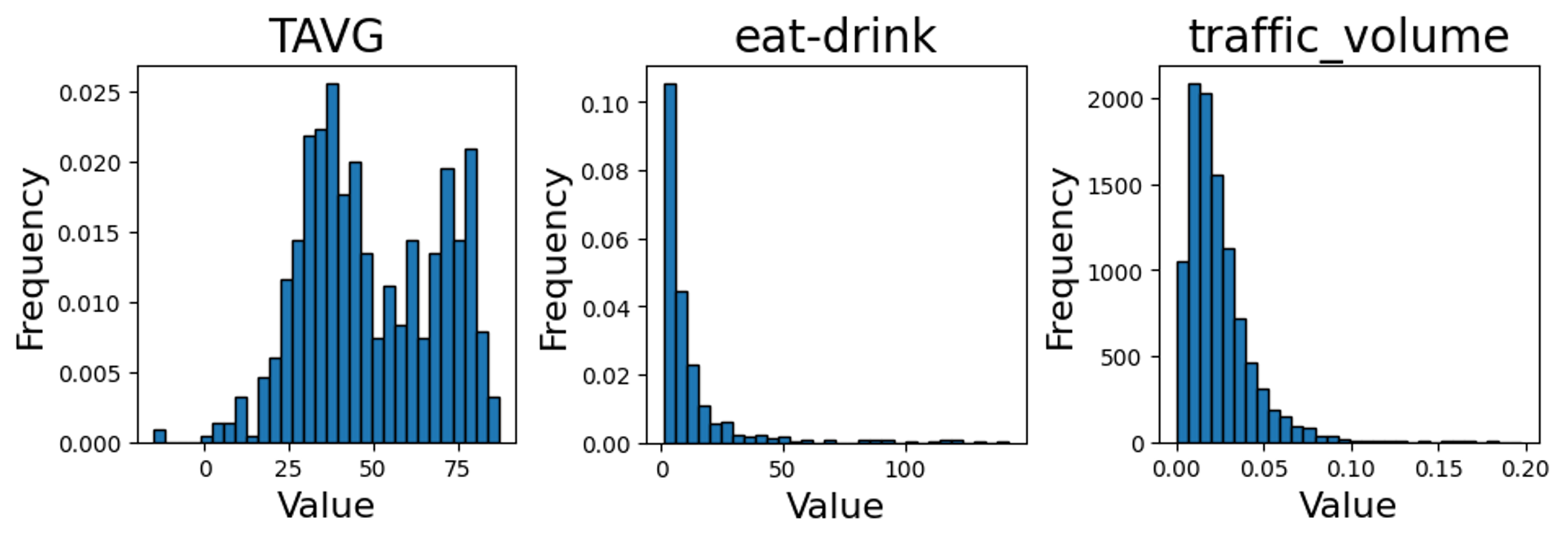}
 \caption{Distribution of traffic volume, average temperature, and eat-drink (Point-of-interests)}
 \label{fig: distribution}
\end{figure}

\begin{figure*}[t]
 \centering
 \includegraphics[width = 1.0\textwidth]{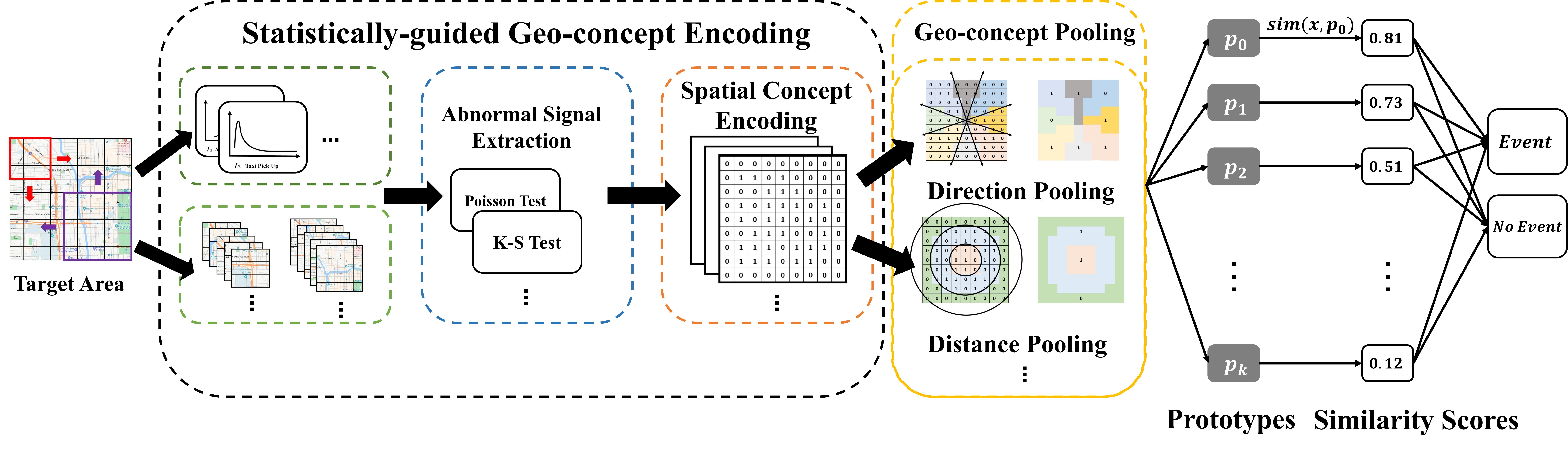}
 \caption{The overall architecture of GeoPro-Net. On the left side, signals are extracted through statistical tests applied to values over features, and then they are mapped into Geo-concepts within the study area. The obtained concepts are further selected via geographical pooling by selected approaches in the middle of the figure. On the right side, a prototype-based framework is integrated to learn the relationships between the occurrence of events and different sets of concept prototypes}
 \label{fig: framework}
\end{figure*}

\begin{definition}
\textbf{Local Significance Test ${\Psi}_L$.} Given a training sample $\mathcal{X} \in \mathbb{R}^{m\times n\times t\times f}$, and a test window $\Lambda \in \mathbb{R}^{d\times d}$, which is a spatial, temporal, or spatiotemporal sub-region of $\mathcal{X}$, ${\Psi}_L(\Lambda, F_i) \rightarrow \{True, False\}$ is a local statistical test function for feature $F_i$ on whether the distribution in $\Lambda$ is \textbf{different} from the distribution of $F_i$ in the entire $\mathcal{X}$ with $\alpha$-level statistical significance with a higher or lower expected value, where $\alpha$ is a hyperparameter. 
\end{definition}

For example, a local significance test ${\Psi}_L$ can be defined as a Poisson likelihood ratio test~\cite{JungInkyung2007Asss} on the number of POIs in a $d\times d$ region of an input sample against the distribution of the number of POIs in all the $d \times d$ sub-regions in the entire sample. An output of ``True'' for the test suggests that this $d\times d$ region might be a local hotspot of POIs near the location to be classified. Similarly, we can define a Global Significance Test as follows. 

\begin{definition}
\textbf{Global Significance Test ${\Psi}_G$.} Given a training sample $\mathcal{X} \in \mathbb{R}^{m\times n\times t\times f}$, and a test window $\Lambda$, which is a spatial, temporal, or spatiotemporal sub-region of $\mathcal{X}$, ${\Psi}_G(\Lambda, F_i) \rightarrow \{True, False\}$ is a global statistical test function for feature $F_i$ on whether the distribution in $\Lambda$ is statistically \textbf{different} from the global distribution of $F_i$ in the entire training set with a higher or lower expected value, with $\alpha$-level significance. 
\end{definition}


Here, the number of dimensions of the window $\Lambda$ depends on the type of feature $F_i$, i.e., one, two, or three dimensional for temporal, spatial, and spatio-temporal features. \underline{The choice of the statistical tests} can be determined by the user based on the nature of the features. Typically, in the urban event forecasting problem, Poisson distributions \cite{1982Pd} are often preferred to estimate the distribution of events, such as counts of taxi-up and traffic volume, which are spatiotemporal features in our dataset. In this paper, we demonstrate the model using a Poisson likelihood ratio test~\cite{vahedian2019predicting} for discrete count features. Other choices of statistical tests such as the student t-test can be used for data commonly assumed to follow a normal distribution. For features without a known or preferred distribution, we use the non-parametric Kolmogorov–Smirnov test as an example\cite{WiesenChristopher2019Ltut}. The Kolmogorov-Smirnov (K-S) test is based on the empirical distribution function. It evaluates the hypothesis that the data comes from a specific distribution against the alternative hypothesis that the data do not follow this specific distribution.
\begin{equation} \label{eqn:ks}
E_N = \frac{n(i)}{N}
\end{equation}
where $n(i)$ is the number of points less than $Y_i$. The value of $Y$ increases with higher $i$. The Kolmogorov-Smirnov test statistic is defined as,
\begin{equation} \label{eqn:ks}
\max_{1\leq i < N}(F(Y_i) -  \frac{i-1}{N}, \frac{i}{N} - F(Y_i)),
\end{equation}
where $Y$ is the value and $F$ is the theoretical cumulative distribution of the distribution being tested. By performing two-tail tests we can test whether a feature value in a sample is significantly higher or significantly lower than the specified baseline distribution. Note the distribution assumptions vary over different problems, and they are user-defined choices that are replaceable in our solution framework.

Now we are ready to define Geo-concepts and the concept convolution operation in GeoPro-Net.
\begin{definition}
\textbf{Geo-concept $\Theta$}. Given an input sample $\mathcal{X} \in \mathbb{R}^{m\times n\times t\times f}$, a Geo-concept $\Theta$ of $\mathcal{X}$ is a tuple $<\Lambda, \Psi, F_i>$, where $\Lambda$ is a sub-region of the spatiotemporal area of $\mathcal{X}$, $\Psi$ is a global or local test on feature $F_i$ in $\Lambda$, which has an output of ``True''. Geo-concept is designed to focus on spatiotemporal interoperability, thus the temporal dimension will be flattened by the mean to reduce the total number of generated concepts.
\end{definition}

Geo-concepts encode statistically significant patterns of input features at both local and global scales in each input sample. We design a Statistically guided Spatial Concept Encoding (SSCE) to extract these concepts and convert the input sample into a multi-channel map tensor that records the positions in the input where these Geo-concepts are valid.



The process of transforming input features into Geo-concept encoding is done through a set of ``Concept Convolution'' operations. From the left part of the black dashed box in Figure ~\ref{fig: framework}, the Statistically-guided Spatial Concept Encoding (SSCE) starts with scanning the samples from the training set across the study area using multiple user-defined window sizes and tests, which is similar to the workflow of the convolutional operation, except that there are no ``parameters'' to train. Instead, users need to specify the choice of statistical tests and the scan window dimensions. 

A concept convolution operation with concept $\Theta = <\Lambda_\Theta, \Psi_\Theta, F_\Theta>$ on an input dataset $\mathcal{X}$ can be defined as follows, where $I()$ is an indicator function with an output of 0 or 1:
\begin{equation}
    ConceptConv_\Theta * \mathcal{X}(\Lambda_\Theta, F_\Theta) = I(\Psi_\Theta(\Lambda_\Theta, F_\Theta) == True)
\end{equation}

 With the use of sliding windows of $\Lambda$, a concept convolution can be applied continuously over the entire input spatiotemporal area. Since multiple window sizes can be used, each input area could be scanned multiple times to capture patterns of various scales. Taking the spatial features $F_s$ of an input sample $\mathcal{X}$ with $m\times n $ spatial dimensions as an example, 
the original input $\mathcal{X}$ will be encoded into a tensor with dimensions in $\mathbb{R}^{m \times n \times f_{s}\times 2\times 4}$, where $\mathbb{R}^{4}$ indicates the global and local tests $\Psi$ on feature $F_i$ in $\Lambda$, which has an output of ``True''. To scan locations on the edges of the input area, padding is applied to these locations with zeros. Spatiotemporal features are encoded in the same way except flatten $\mathbb{R}^t$ by averaging overtime in the beginning.

Formally, the encoder $E$ can be defined as a function over input $\mathcal{X} \in \mathbb{R}^{m\times n\times (f_{st}+f_s)}$: $E(\mathcal{X})\rightarrow C\in \{0,1\}^{m\times n \times (f_{st}+f_s) \times \omega\times 4}$ for $F_S$ and $F_{ST}$, which generates a binary concept map tensor $C$, 
$\omega$ is the number of different concept convolution window sizes for each feature. The last dimension of $C$ indicates four combinations of global vs local tests and higher or lower expected values in the test. Temporal features are encoded in the same way but removing spatial dimensions $\mathbb{R}^{m \times n}$.

The use of multiple concept window sizes could capture patterns of varying scales. These concepts may overlap and contain redundant information as a result. We design a Channel Fusion Operation and Multi-scale geo-pooling on extracted raw Geo-concepts to address these issues, which is detailed in the next subsection.

\subsection{Geo-Concept Aggregation}
\textbf{Geo-concept Channel Fusion} In GeoPro-Net, multiple sets of concepts are generated on each input sample by corresponding scanning windows, which increases the amount of extracted concepts and poses challenges for interpretation. To address these issues, we design a Channel Fusion Operation on the different channels of the concept encoding $C$ 
to consolidate and refine the concept encoding into more concise but still interpretable representations. Especially, for each window size of the same test (i.e., channels), we learn a weight matrix $\mathbf{w}_i$ with the same dimensions as each raw concept encoding channel 
and combine the channels as a weighted some of the features. Formally,
\begin{equation}
    C'_{F, \Theta} = \Sigma_{i=1}^\omega C_{F, \Theta, \Lambda_i} * \mathbf{w}_i, \text{where}\  \Sigma_{i=1}^\omega \mathbf{w}_i = \mathbf{1^{m\times n}}
\end{equation}
Here $C_{F, \Theta,\Lambda_i}$ represents the encoded input channel for feature $F$ convoluted by concept $\Theta$ with a scanning window of $\Lambda_i$, which is combined into a single channel for feature $F$ and $\Theta$, denoted as $C'_{F, \Theta}$, and $*$ is the Hadamard product. 
Eventually, we obtain a fused concept tensor $C' \in \mathbb{R}^{(f_s+f_{st}) \times m \times n \times 4}$. 


\begin{figure}[ht] \hspace*{0cm}
 \centering
 \includegraphics[width = 0.45\textwidth]{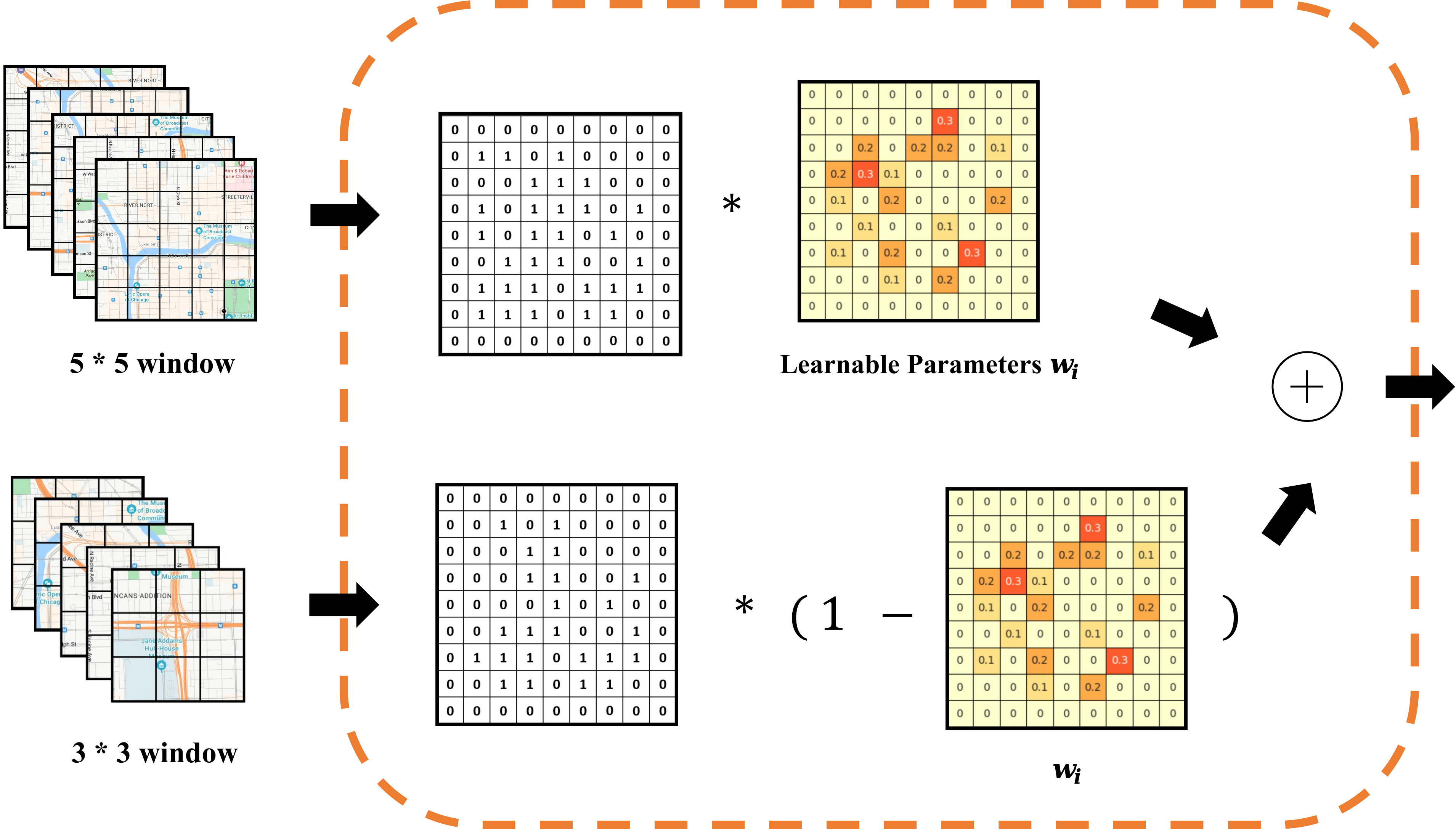}
 \caption{Geo-concept Channel Fusion. Encoded concepts are fused by weighted element-wise summation}
 \label{fig: fusion}
\end{figure}

\textbf{Geo-concept Pooling.} To further enhance the semantic representation of extracted concepts and reduce the encoding dimensions, we design a Geo-concept pooling layer with user-defined pooling strategies. 
The dashed yellow box in Figure \ref{fig: framework} illustrates the pooling process. Depending on the target problem, the study area can be partitioned into sub-regions, and the mean or average of encoding $C'$ is calculated from each pooled sub-region. In this problem, urban events are highly associated with nearby conditions \cite{ToblerW.R.1970ACMS}. Therefore, we partition the study area into near, middle, and far sub-regions based on the Euclidean distance between each grid cell and the target location at the center of the sample. Meanwhile, we partition the study area into different directions based on the degree between the coordinates of each location and the central target location. Finally, the concept tensor is reshaped to $C' \in \mathbb{R}^{f' \times q \times 4}$, where $f' = (f_s + f_{st})\times q + f_t$ and $q$ represents the number of different pooling regions and $4$ indicates the statistical test types. 
The pooling layer maintains the interpretability of the model. A positive value in a pooled region indicates the presence of at least one statistically significant pattern, while the magnitude of the pooled values indicates the overall strength of such presences. For example, an average-pooled value of 0.2 over a sub-region with 10 grid cells indicates at least two instances of statistically significant patterns in this sub-region.

\subsection{Concept Prototype Layer} In GeoPro-Net, the prototype layer is illustrated on the right part of Figure  \ref{fig: framework}. The prototype layer is a learnable parameter $P' \in \mathbb{R}^{K \times f' * q * 4}$, where each prototype is a vector and $K$ is the number of prototypes in our model. $K$ is a hyperparameter and can be adjusted. The concept prototypes share the same dimension as pooled Geo-concept encoding and will be learned to be close to a subset of pooled concepts during training. The prototype learning is related to case-based classifications \cite{PRIEBECareyE2003Cucc}\cite{NEURIPS2019_adf7ee2d}\cite{BienJacob2011PSFI}, and $K$ representative prototypes are learned to be compared with unseen testing samples. Given a Geo-concept encoded testing sample $C \in \mathbb{R}^{f' \times q \times 4}$, the squared $L^2$ distances to a set of prototypes $P' \in \mathbb{R}^{K \times f' \times q \times 4}$ are inverted into $K$ similarity scores, 
\begin{equation} \label{eqn:ks}
Sim(C, p_k) = \frac{1}{||C - p_k||^2_2}
\end{equation}
where $p$ is the $k^{\text{th}}$ learned prototype vector. Lastly, all similarity scores are connected to classifications, distinguishing the occurrence of an event versus no event, through a fully connected layer. The coefficients of connections indicate the linear correlation between prototypes and their corresponding class.

\subsection{Optimization}

The training of GeoPro-Net requires diversifying and disambiguating the learned prototypes to ensure representability and interpretability. This section outlines three regularization terms and a prototype projection process to achieve this goal. 
Related prototype-based models \cite{NEURIPS2019_adf7ee2d}\cite{YaoMing2019IaSS} demonstrate the effectiveness of diversifying and disambiguating prototypes by using such regularization. Specifically, the $Dlv$ (Diversity) term encourages model learning different prototypes, $Sep$ (Separation) term pushes encoded concepts vector to stay away from the prototypes not of its class, and $Clst$ (Cluster) encourages encoded concepts vector to be close to the prototype of its class. 

\begin{equation} \label{eqn:ks}
Dilv = \min_{p_i \in P, p_j \in P} - \frac{1}{|P|^2} ||p_i - p_j||^2_2,
\end{equation}
\begin{equation} \label{eqn:ks}
Sep = - \frac{1}{n} \sum_{i=1}^n \min_{p_j \notin P_{y_i}} \min_{z\in(f(x_i))} || z - p_j||^2_2,
\end{equation}
\begin{equation} \label{eqn:ks}
Clst = \frac{1}{n} \sum_{i=1}^n \min_{p_j \in P_{y_i}} \min_{z\in(f(x_i))} || z - p_j||^2_2.
\end{equation}

This approach simplifies interpretation by presenting only an appropriate number of significant concepts for each prototype, making the information more comprehensible and interpretable.
The overall loss function for training our model is defined as minimizing
\begin{equation} \label{eqn:ks}
\mathcal{L} = \frac{1}{n} \sum_{i=1}^n \text{CrsEnt}(g(x_i), y_i) + \lambda_1 Dlv + \lambda_2 Sep + \lambda_3 Clst
\end{equation}
where the $g()$ is the learned model, and CrsEnt represents the cross-entropy loss, 

\textbf{Complexity Analysis} GeoPro-Net achieves practical scalability with highly efficient inference. The initial preprocessing involves calculating local and global feature distributions via Local Significance Test $\Psi_L$ and Global Significance Test $\Psi_G$ across multiple window sizes $\omega$ over a spatial grid of size $m\times n$, with complexity $O(\omega \times m \times n \times t \times f)$ per feature $F_i$. Though computationally costly during training, the historical baseline only needs to be calculated once and can be reused during inference, making Concept Convolution more efficient. Once training data is encoded, geo-concepts achieve through Channel Fusion and Geo-concept Pooling, which refines the concept map $C$ by consolidating channels $\theta'=\sum{\omega\times C(F, \theta, \Lambda)}$ and aggregating pooled sub-regions, resulting in complexity of $O(f' \times m \times n)$. The resulting Geo-concept tensor, a list of vectors, enables efficient prototype matching. Geo-concept encodings are matched with learned prototypes P, yielding in an efficient complexity of $O(K \times f' \times q)$.

\textbf{Concept Prototype Projection} is a post-training process. To interpret the learned concept prototypes, they are projected back into encoded Geo-concepts from the training set. Each prototype represents a real scenario by demonstrating a sample of encoded concepts. Specifically, we find the sample of encoded concepts with maximum similarity score to every prototype. This set of extracted concepts can be visualized in an easily digestible manner by listing only the most significant ones

\section{Experiments}

\begin{table*}[ht]
\fontsize{9}{11}\selectfont
\centering
\caption{Levels of interpretability (IT)}
\setlength{\tabcolsep}{2mm}{
\begin{tabular}{c|c|c|c|c|c|c|c|c|c|c}\toprule
\multirow{1}{*}{} &  \textbf{ConvLSTM} & \textbf{DCRNN} & \textbf{GSNet}  &  \textbf{HeteroConvLSTM}  & \textbf{DSTPP}  & \textbf{ViT}  & \textbf{ASTGCN}  &  \textbf{ProtoPNet} &  \textbf{GeoPro-Net}  \\\midrule

		IT & Blackbox & Blackbox & Blackbox  & Blackbox & Blackbox & Part-Interp. & Part-Interp. & Interp. & Interp. \\
		\hline

\bottomrule
\end{tabular}}
\label{table: levels}
\end{table*}

We perform comprehensive experiments using four real-world traffic accident and crime datasets from Chicago{\footnote{\tiny{https://data.cityofchicago.org/Transportation/Traffic-Crashes-Crashes/85ca-t3if}} and New York City{\footnote{\tiny{https://opendata.cityofnewyork.us/}}. The experiments and case studies demonstrate that GeoPro-Net achieves better interpretability than interpretable competitors and higher performance than Blackbox baselines.

\begin{table*}[t]
\begin{center}
\begin{threeparttable}[b]
\caption{Performance Comparison}
\label{tab:compare}
\begin{small}
\begin{sc}
\begin{tabular}
{p{2.3cm}p{1.1cm}p{1.1cm}p{1.1cm}p{1.1cm}p{1.1cm}p{1.1cm}p{1.1cm}p{1.1cm}p{1.1cm}p{1.1cm}}
\toprule

\multirow{1}{*}{\thead{\textbf{New York City}}} &
\multicolumn{5}{c}{\thead{Crime}} &
\multicolumn{5}{c}{\thead{Accident}} \\
\cmidrule(lr){2-6}
\cmidrule(lr){7-11}

& \footnotesize{CrsEnt} & \footnotesize{ACC} & \tiny{Precision} & \footnotesize{Recall} & \footnotesize{F1} & \footnotesize{CrsEnt} & \footnotesize{ACC} & \tiny{Precision} & \footnotesize{Recall} & \footnotesize{F1} \\
\midrule
ConvLSTM   & .62$\pm$2.6\tiny\textperthousand  & .76$\pm$0.8\tiny\textperthousand & .51$\pm$2.5\tiny\textperthousand & .19$\pm$0.8\tiny\textperthousand & .33$\pm$4.4\tiny\textperthousand & .60$\pm$1.6\tiny\textperthousand & .73$\pm$7.7\tiny\textperthousand & .40$\pm$3.8\tiny\textperthousand & .19$\pm$1.0\tiny\textperthousand  & .25$\pm$0.9\tiny\textperthousand \\
DCRNN   & .53$\pm$1.6\tiny\textperthousand  & .77$\pm$3.1\tiny\textperthousand & .66$\pm$1.1\tiny\textperthousand & .27$\pm$0.6\tiny\textperthousand & .40$\pm$1.8\tiny\textperthousand & .56$\pm$1.8\tiny\textperthousand & .75$\pm$4.1\tiny\textperthousand & .49$\pm$2.0\tiny\textperthousand & .26$\pm$0.7\tiny\textperthousand  & .36$\pm$1.1\tiny\textperthousand \\
\tiny HeteroConvLSTM   & .52$\pm$2.6\tiny\textperthousand  & .78$\pm$3.5\tiny\textperthousand & .65$\pm$1.6\tiny\textperthousand & \textbf{.28}$\pm$1.0\tiny\textperthousand & .42$\pm$1.1\tiny\textperthousand & .55$\pm$1.5\tiny\textperthousand & .74$\pm$2.0\tiny\textperthousand & \textbf{.50}$\pm$2.3\tiny\textperthousand & .27$\pm$0.6\tiny\textperthousand  & .35$\pm$1.3\tiny\textperthousand \\
DSTPP   & .57$\pm$0.7\tiny\textperthousand  & .76$\pm$1.1\tiny\textperthousand & .60$\pm$0.5\tiny\textperthousand & .22$\pm$0.6\tiny\textperthousand & .36$\pm$0.7\tiny\textperthousand & .58$\pm$0.9\tiny\textperthousand & .72$\pm$0.8\tiny\textperthousand & .44$\pm$2.0\tiny\textperthousand & .25$\pm$1.2\tiny\textperthousand  & .29$\pm$1.7\tiny\textperthousand \\
ViT   & .54$\pm$1.3\tiny\textperthousand  & .77$\pm$2.6\tiny\textperthousand & .65$\pm$1.9\tiny\textperthousand & .26$\pm$0.5\tiny\textperthousand & .39$\pm$1.4\tiny\textperthousand & .58$\pm$1.5\tiny\textperthousand & .74$\pm$3.7\tiny\textperthousand & .46$\pm$2.7\tiny\textperthousand & .27$\pm$0.9\tiny\textperthousand  & .35$\pm$0.3\tiny\textperthousand \\
GSNet   & .56$\pm$0.3\tiny\textperthousand  & .77$\pm$2.4\tiny\textperthousand & .53$\pm$1.4\tiny\textperthousand & .25$\pm$0.5\tiny\textperthousand & .35$\pm$0.7\tiny\textperthousand & .57$\pm$1.4\tiny\textperthousand & .73$\pm$0.7\tiny\textperthousand & .46$\pm$1.8\tiny\textperthousand & .24$\pm$0.8\tiny\textperthousand  & .31$\pm$1.3\tiny\textperthousand \\
AGL-STAN   & .56$\pm$0.9\tiny\textperthousand  & .77$\pm$1.1\tiny\textperthousand & .56$\pm$0.5\tiny\textperthousand & .25$\pm$0.8\tiny\textperthousand & .37$\pm$1.1\tiny\textperthousand & .57$\pm$0.3\tiny\textperthousand & .74$\pm$1.1\tiny\textperthousand & .47$\pm$0.4\tiny\textperthousand & .26$\pm$0.9\tiny\textperthousand  & .33$\pm$1.7\tiny\textperthousand \\
ASTGCN   & .57$\pm$0.3\tiny\textperthousand  & .76$\pm$0.4\tiny\textperthousand & .55$\pm$1.0\tiny\textperthousand & .26$\pm$0.4\tiny\textperthousand & .37$\pm$0.6\tiny\textperthousand & .57$\pm$0.3\tiny\textperthousand & .74$\pm$0.9\tiny\textperthousand & .48$\pm$0.5\tiny\textperthousand & .27$\pm$0.4\tiny\textperthousand  & .34$\pm$0.6\tiny\textperthousand \\
ProtoPNet   & .58$\pm$0.0\tiny\textperthousand  & .77$\pm$0.4\tiny\textperthousand & .59$\pm$0.1\tiny\textperthousand & .22$\pm$0.2\tiny\textperthousand & .36$\pm$0.3\tiny\textperthousand & .59$\pm$0.9\tiny\textperthousand & .73$\pm$0.5\tiny\textperthousand & .43$\pm$0.1\tiny\textperthousand & .20$\pm$0.1\tiny\textperthousand  & .28$\pm$0.3\tiny\textperthousand \\
GeoPro-Net*   & .53$\pm$0.5\tiny\textperthousand  & .77$\pm$0.4\tiny\textperthousand & .64$\pm$0.4\tiny\textperthousand & .24$\pm$0.0\tiny\textperthousand & .39$\pm$0.3\tiny\textperthousand & .57$\pm$0.6\tiny\textperthousand & .74$\pm$0.7\tiny\textperthousand & .47$\pm$0.6\tiny\textperthousand & .26$\pm$0.2\tiny\textperthousand  & .33$\pm$0.3\tiny\textperthousand \\
GeoPro-Net   & \textbf{.51}$\pm$0.3\tiny\textperthousand  & \textbf{.79}$\pm$0.1\tiny\textperthousand & \textbf{.68}$\pm$0.4\tiny\textperthousand & .27$\pm$0.2\tiny\textperthousand & \textbf{.43}$\pm$0.2\tiny\textperthousand & \textbf{.54}$\pm$0.2\tiny\textperthousand & \textbf{.76}$\pm$0.2\tiny\textperthousand & \textbf{.50}$\pm$0.6\tiny\textperthousand & \textbf{.28}$\pm$0.2\tiny\textperthousand  & \textbf{.37}$\pm$0.4\tiny\textperthousand \\

\hline

\multirow{1}{*}{\thead{\textbf{Chicago}}} &
\multicolumn{5}{c}{\thead{Crime}} &
\multicolumn{5}{c}{\thead{Accident}} \\
\cmidrule(lr){2-6}
\cmidrule(lr){7-11}

& \footnotesize{CrsEnt} & \footnotesize{ACC} & \tiny{Precision} & \footnotesize{Recall} & \footnotesize{F1} & \footnotesize{CrsEnt} & \footnotesize{ACC} & \tiny{Precision} & \footnotesize{Recall} & \footnotesize{F1} \\
\midrule
ConvLSTM   & .59$\pm$4.4\tiny\textperthousand  & .75$\pm$6.3\tiny\textperthousand & .46$\pm$7.2\tiny\textperthousand & .17$\pm$0.6\tiny\textperthousand & .21$\pm$0.8\tiny\textperthousand & .61$\pm$5.6\tiny\textperthousand & .72$\pm$3.1\tiny\textperthousand & .30$\pm$1.5\tiny\textperthousand & .18$\pm$1.6\tiny\textperthousand  & .23$\pm$1.1\tiny\textperthousand \\
DCRNN   & .53$\pm$1.4\tiny\textperthousand  & .78$\pm$4.1\tiny\textperthousand & .52$\pm$2.3\tiny\textperthousand & .23$\pm$0.9\tiny\textperthousand & .25$\pm$0.6\tiny\textperthousand & .54$\pm$2.9\tiny\textperthousand & .74$\pm$4.6\tiny\textperthousand & .39$\pm$2.7\tiny\textperthousand & .27$\pm$1.1\tiny\textperthousand  & \textbf{.34}$\pm$1.6\tiny\textperthousand \\
\tiny HeteroConvLSTM   & .50$\pm$1.4\tiny\textperthousand & .79$\pm$1.2\tiny\textperthousand & \textbf{.57}$\pm$2.0\tiny\textperthousand & .24$\pm$0.4\tiny\textperthousand & .28$\pm$1.3\tiny\textperthousand & .54$\pm$1.4\tiny\textperthousand & .73$\pm$2.0\tiny\textperthousand & .40$\pm$4.4\tiny\textperthousand & .29$\pm$1.4\tiny\textperthousand  & .32$\pm$2.2\tiny\textperthousand \\
DSTPP   & .55$\pm$1.9\tiny\textperthousand  & .76$\pm$1.8\tiny\textperthousand & .53$\pm$0.8\tiny\textperthousand & .21$\pm$0.9\tiny\textperthousand & .24$\pm$0.8\tiny\textperthousand & .58$\pm$2.2\tiny\textperthousand & .72$\pm$2.9\tiny\textperthousand & .36$\pm$1.9\tiny\textperthousand & .21$\pm$1.2\tiny\textperthousand  & .27$\pm$1.5\tiny\textperthousand \\
ViT   & .53$\pm$0.9\tiny\textperthousand  & .77$\pm$2.7\tiny\textperthousand & .54$\pm$1.7\tiny\textperthousand & .23$\pm$1.3\tiny\textperthousand & .24$\pm$1.1\tiny\textperthousand & .55$\pm$2.8\tiny\textperthousand & .73$\pm$3.2\tiny\textperthousand & .39$\pm$2.3\tiny\textperthousand & .26$\pm$1.6\tiny\textperthousand  & .33$\pm$0.9\tiny\textperthousand \\
GSNet   & .51$\pm$1.6\tiny\textperthousand  & .78$\pm$1.5\tiny\textperthousand & .54$\pm$0.9\tiny\textperthousand & .23$\pm$0.7\tiny\textperthousand & .28$\pm$0.5\tiny\textperthousand & .56$\pm$2.1\tiny\textperthousand & .73$\pm$3.5\tiny\textperthousand & .36$\pm$1.4\tiny\textperthousand & .22$\pm$1.0\tiny\textperthousand  & .28$\pm$1.3\tiny\textperthousand \\
AGL-STAN   & .53$\pm$1.2\tiny\textperthousand  & .78$\pm$1.8\tiny\textperthousand & .52$\pm$0.5\tiny\textperthousand & .24$\pm$0.9\tiny\textperthousand & .27$\pm$1.1\tiny\textperthousand & .56$\pm$1.1\tiny\textperthousand & .72$\pm$0.3\tiny\textperthousand & .35$\pm$2.1\tiny\textperthousand & .21$\pm$1.2\tiny\textperthousand  & .28$\pm$0.8\tiny\textperthousand \\
ASTGCN   & .53$\pm$1.4\tiny\textperthousand  & .78$\pm$2.4\tiny\textperthousand & .53$\pm$0.8\tiny\textperthousand & .23$\pm$0.5\tiny\textperthousand & .26$\pm$0.8\tiny\textperthousand & .57$\pm$0.9\tiny\textperthousand & .73$\pm$0.3\tiny\textperthousand & .34$\pm$1.3\tiny\textperthousand & .23$\pm$0.2\tiny\textperthousand  & .30$\pm$0.7\tiny\textperthousand \\
ProtoPNet   & .56$\pm$0.5\tiny\textperthousand  & .78$\pm$0.7\tiny\textperthousand & .50$\pm$0.4\tiny\textperthousand & .19$\pm$0.2\tiny\textperthousand & .23$\pm$0.2\tiny\textperthousand & .59$\pm$0.5\tiny\textperthousand & .72$\pm$0.6\tiny\textperthousand & .38$\pm$0.3\tiny\textperthousand & .21$\pm$0.3\tiny\textperthousand  & .27$\pm$0.4\tiny\textperthousand \\
GeoPro-Net*   & .53$\pm$0.8\tiny\textperthousand  & .78$\pm$0.1\tiny\textperthousand & .50$\pm$0.4\tiny\textperthousand & .21$\pm$0.2\tiny\textperthousand & .24$\pm$0.2\tiny\textperthousand & .56$\pm$0.5\tiny\textperthousand & .73$\pm$0.6\tiny\textperthousand & .36$\pm$0.3\tiny\textperthousand & .24$\pm$0.2\tiny\textperthousand  & .29$\pm$0.3\tiny\textperthousand \\
GeoPro-Net   & \textbf{.49}$\pm$0.3\tiny\textperthousand  & \textbf{.80}$\pm$0.7\tiny\textperthousand & .55$\pm$0.8\tiny\textperthousand & \textbf{.25}$\pm$0.4\tiny\textperthousand & \textbf{.30}$\pm$0.3\tiny\textperthousand & \textbf{.53}$\pm$0.5\tiny\textperthousand & \textbf{.76}$\pm$0.5\tiny\textperthousand & \textbf{.41}$\pm$0.1\tiny\textperthousand & \textbf{.29}$\pm$0.3\tiny\textperthousand  & \textbf{.34}$\pm$0.3\tiny\textperthousand \\

\bottomrule
\end{tabular}
\begin{tablenotes}
    \item \textperthousand: $\times10^{-3}$
    \item *: replace two concept convolution windows by a single $3 \times 3$ window
\end{tablenotes}
\end{sc}
\end{small}
\end{threeparttable}
\end{center}

\label{table: comparision}
\end{table*}

\subsection{Data.} For the Chicago dataset, the time frame spans from 2019 to 2021, with the initial 18 months serving as the training set, while the entirety of 2021 as the testing set. The area of Chicago is partitioned by $500$ m $\times$ $500$ m square cells and converted to a grid with the size of $64 \times 80$. In the New York City dataset, we collect data from 2020 is used for training, while the year of 2021 is employed for testing. The area of New York City is partitioned by $550$ m $\times$ $550$ m square cells and converted to a grid with the size of $68 \times 55$. For each location, features are extracted from neighboring locations in squared windows with size $9 \times 9$. Note that changing the resolution of grids might influence the model performance, and we choose this granularity setting to leverage between problem difficulty and computational efficiency. All baselines are experimented with the same settings. The experiment results in New York City can be found in Appendix B.

\subsection{Evaluation Goals}
We wish to answer the following questions in the experiments: (1) Does the proposed GeoPro-Net exhibit comparable performance to other Blackbox baselines? (2) Does GeoPro-Net archive better interpretability than other state-of-arts explainers? (3) How do parameter settings influence model performance and interpretability? (4) Can GeoPro-Net capture diverse patterns and provide interpretation in the spatially heterogeneous areas?

\subsection{Parameter Configurations}
The proposed method is trained by minimizing the mean squared error by using the Adam optimizer \cite{KingmaDiederikP2014AAMf} with settings $\alpha=10^{-3}$, $\beta_1=0.9$, $\beta_2=0.999$, and $\epsilon=10^{-8}$. An early stopping mechanism is employed while training models, and the training process is terminated if the validating loss stops decreasing for 5 consecutive epochs. Significance levels $\alpha$ of the statistical tests are set as $0.05$.

\subsubsection{Platform}
We run the experiments on High-Performance Computer System with Intel Xeon E5 2.4 GHz and 256 GB of Memory. We use a GPU node with Nvidia Tesla V100 Accelerator Cards with the support of Pytorch library \cite{NEURIPS2019_9015} to train the deep learning models.

\begin{figure*}[t]
 \centering
 \includegraphics[width = 0.8\textwidth]{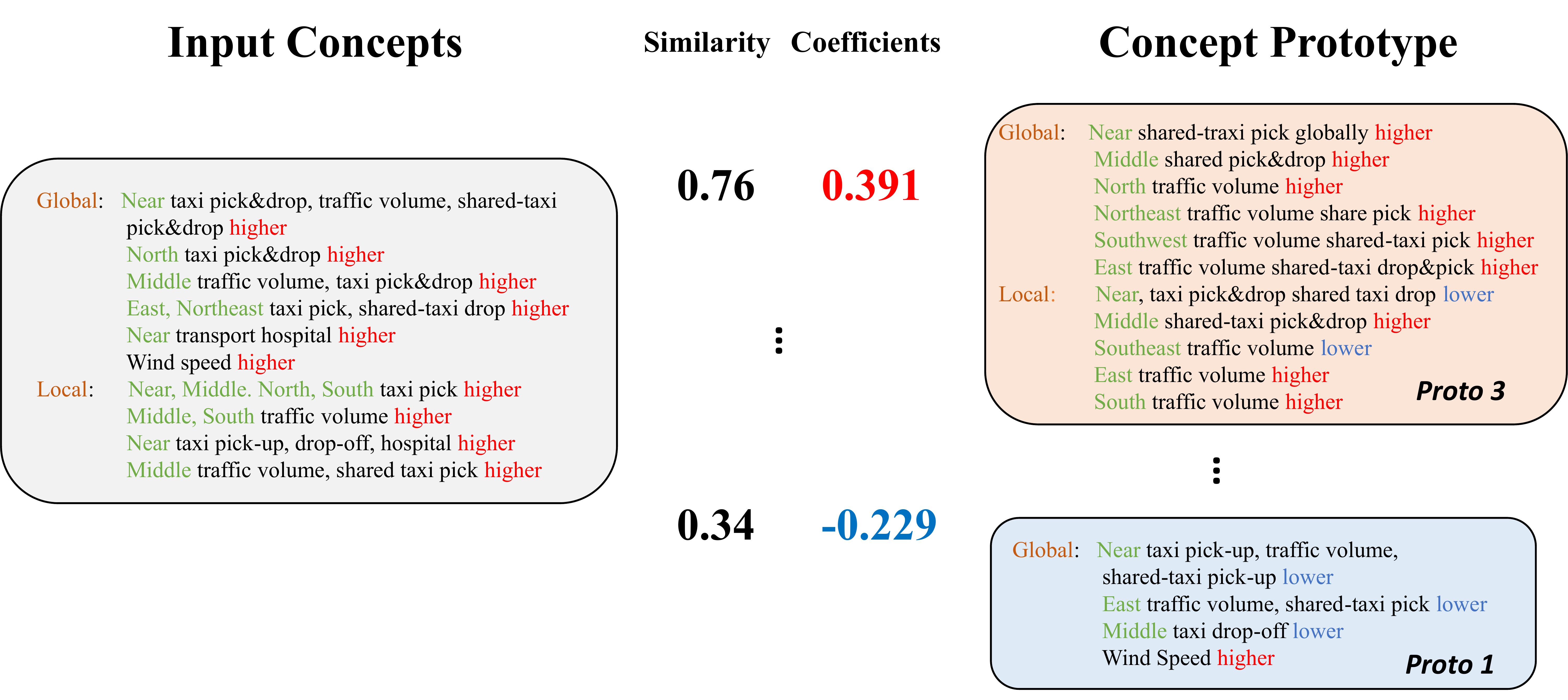}
 \caption{Grey boxes represent the encoded concepts for this given case, and red and blue boxes represent learned concept prototypes with positive and negative correlations to the occurrence of events.}
 \label{geo: chicago geopro2}
\end{figure*}

\subsection{Baselines}
We compare the proposed method with the following baselines.

\textbf{(1) ConvLSTM} \cite{ShiXingjian2015CLNA} is an early spatiotemporal deep model with convolution layers for precipitation nowcasting. 

\textbf{(2) DCRNN} \cite{li2018diffusion} is an advanced traffic forecasting model which captures the spatial dependency using bidirectional random walks on the graph and the temporal dependency using an encoder-decoder architecture.

\textbf{(3) Hetero-ConvLSTM} \cite{yuan2018hetero} is an advanced deep learning framework to address spatial heterogeneity in the traffic accident forecasting problem.

\textbf{(4) DSTPP} \cite{yuan2023DSTPP} Deep Spatio-temporal point process (DSTPP) is a stochastic collection of events accompanied with time and space. We assign predictions of closest point to each locations to evaluate model perforamnces.

\textbf{(5) ViT} \cite{dosovitskiy2020vit} The Vision Transformer (ViT) applies transformer models to image recognition and can be interpretable through attention map visualizations. 

\textbf{(6) GSNet} \cite{Wang2021gsnet} is a recent deep-learning method to learn geographical and semantic aspects for traffic accident prediction.

\textbf{(7) ProtoPNet} \cite{NEURIPS2019_adf7ee2d}is an intrinsically interpretable model for image classification, and it works by classifying images based on comparing the similarity between part of the input images and learned prototypes.

\textbf{(8) AGL-STAN} \cite{SunMingjie2022SANf} is an attention-based model for efficiently capturing complex spatial-temporal correlations of urban crimes with higher prediction accuracy.

\textbf{(9) ASTGCN} \cite{guo2019attention} is an attention-based spatiotemporal graph neural network for traffic problems. It captures the spatiotemporal dynamics of traffic data by using an attention mechanism, which is partially interpretable via its attention map.

\subsection{Categorization of Model Interpretability}

Based on the interpretability of each model, We categorize different methods into the Blackbox model, partially interpretable model (Part-Interp), and intrinsically interpretable model (Interp) in Table. \ref{table: levels}. ConvLSTM, DCRNN, GSnet, and Hetero-ConvLSTM, relying on complicated deep neural networks, are categorized as Blackbox competitors designed to minimize prediction errors without prioritizing interpretability. On the other hand, ASTGCN, an attention-based post-hoc explainer, falls into the partially interpretable category as it is limited to the analysis of attention maps but cannot explain the correlations between features and events. ProtoPNet is a prototype-based intrinsically interpretable model, capable of demonstrating positive or negative correlations between the occurrence of events and learned prototypes. Nevertheless, ProtoPNet fails to provide deeper insights into causality between the occurrence of events and related factors. Notably, Our proposed method, GeoPro-Net, is an intrinsically interpretable model to learn spatiotemporal features through statistically-guided Geo-prototyping. Detailed comparisons and visualizations of interpretability for each method are elaborated in case studies.

\subsection{Evaluation on Model Performance}


In table \ref{tab:compare}, we can observe that GeoPro-Net achieves better performance compared with its Blackbox competitors in most measurements from four datasets. Specifically, ConvLSTM performs worst due to its preference for precipitation now-casting problems rather than urban event datasets used in the experiments. The state-of-art Blackbox models such as DCRNN and Hetero-ConvLSTM benefit from complex network design and perform well compared to interpretable baselines. ProtoPNet and ViT, designed for image classification problems, struggles with handling the spatiotemporal features, resulting in comparatively worse performance. GeoPro-Net, with its simple prototype-based network structure, demonstrates stable performance and outperforms other baselines. 
We conduct an ablation study on the channel fusion part. GeoPro-Net with two concept convolution windows performs better than the model with one window. 

\begin{table}[t]
\begin{center}
\begin{threeparttable}[b]
\caption{Ablation Study}
\label{tab:ablation}
\begin{small}
\begin{sc}
\begin{tabular}{p{2.0cm}p{0.9cm}p{0.9cm}p{0.9cm}p{0.9cm}p{0.9cm}}
\toprule

\multirow{1}{*}{\thead{\textbf{Chicago}}} &
\multicolumn{5}{c}{\thead{Crime}} \\
\cmidrule(lr){2-6}

& \footnotesize{CrsEnt} & \footnotesize{ACC} & \tiny{Precision} & \footnotesize{Recall} & \footnotesize{F1} \\
\midrule
No Pooling & .67$\pm$2.5\tiny\textperthousand  & .71$\pm$1.4\tiny\textperthousand & .39$\pm$2.4\tiny\textperthousand & .17$\pm$0.5\tiny\textperthousand & .23$\pm$1.2\tiny\textperthousand  \\
Max Pooling & .51$\pm$0.8\tiny\textperthousand  & \textbf{.80}$\pm$0.7\tiny\textperthousand & .52$\pm$0.8\tiny\textperthousand & .23$\pm$0.5\tiny\textperthousand & .28$\pm$1.2\tiny\textperthousand  \\
Spatial* & \textbf{.49}$\pm$0.3\tiny\textperthousand  & \textbf{.80}$\pm$0.7\tiny\textperthousand & \textbf{.55}$\pm$0.8\tiny\textperthousand & \textbf{.25}$\pm$0.4\tiny\textperthousand & \textbf{.30}$\pm$0.3\tiny\textperthousand  \\

\multirow{1}{*}{\thead{ }} &
\multicolumn{5}{c}{\thead{Accident}} \\
\cmidrule(lr){2-6}

& \footnotesize{CrsEnt} & \footnotesize{ACC} & \tiny{Precision} & \footnotesize{Recall} & \footnotesize{F1} \\
\midrule
No Pooling & .69$\pm$3.7\tiny\textperthousand  & .67$\pm$3.6\tiny\textperthousand & .28$\pm$0.4\tiny\textperthousand & .21$\pm$0.6\tiny\textperthousand & .26$\pm$0.8\tiny\textperthousand \\
Max Pooling & .55$\pm$0.5\tiny\textperthousand  & .73$\pm$1.1\tiny\textperthousand & .39$\pm$0.6\tiny\textperthousand & .28$\pm$0.2\tiny\textperthousand & .33$\pm$0.5\tiny\textperthousand \\
Spatial* & \textbf{.53}$\pm$0.5\tiny\textperthousand  & \textbf{.76}$\pm$0.5\tiny\textperthousand & \textbf{.41}$\pm$0.1\tiny\textperthousand & \textbf{.29}$\pm$0.3\tiny\textperthousand & \textbf{.34}$\pm$0.3\tiny\textperthousand  \\
\bottomrule
\end{tabular}
\begin{tablenotes}
    \item *: Spaital Concept Pooling
    \item $'$: $\times10^{-3}$
\end{tablenotes}
\end{sc}
\end{small}
\end{threeparttable}
\end{center}

\label{table: ablation2}
\end{table}

\subsection{Ablation Study}

We examine the effects of removing spatial pooling or replacing with max pooling in the GeoPro-Net. Table. 2 shows that GeoPro-Net achieves optimal performance with spatial concept pooling. Replacing spatial pooling by max pooing slightly lowers the performance and loses the capability to interpret based on geographic information. Removing the pooling layer results in a dramatic decrease in its performance due to noisy information.

\begin{figure}
\centering
\begin{minipage}[c]{0.45\textwidth}
\centering\includegraphics[width=1\textwidth]{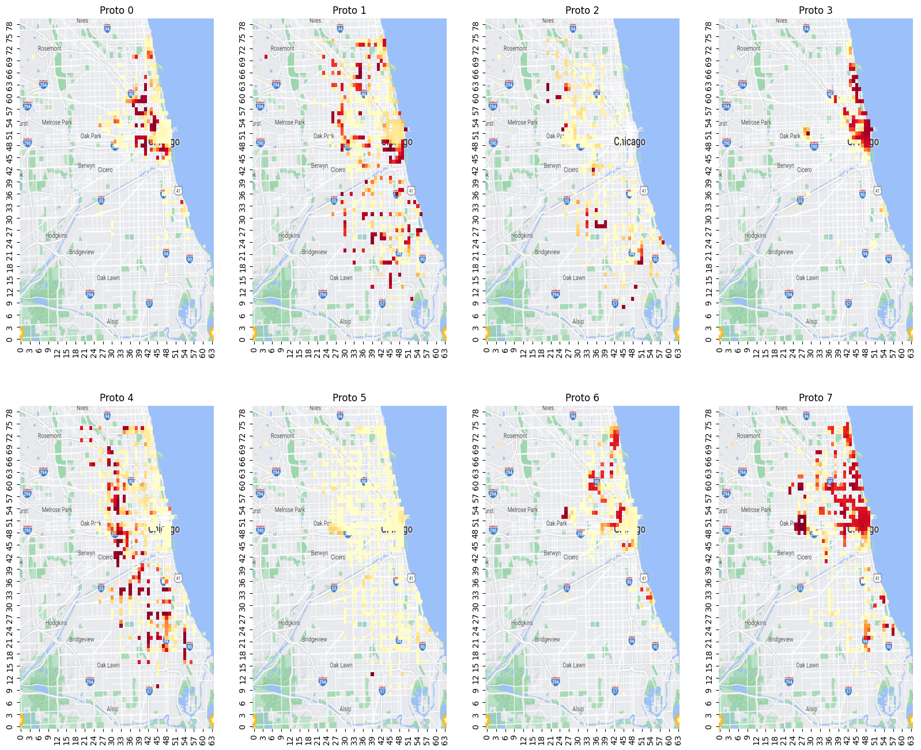}
   \caption{Chicago: Mapping Similar Samples to Every Prototype over Space}
    \label{geo: chicago geopro3}
\end{minipage}
\begin{minipage}[c]{0.45\textwidth}
\centering\includegraphics[width=1\textwidth]{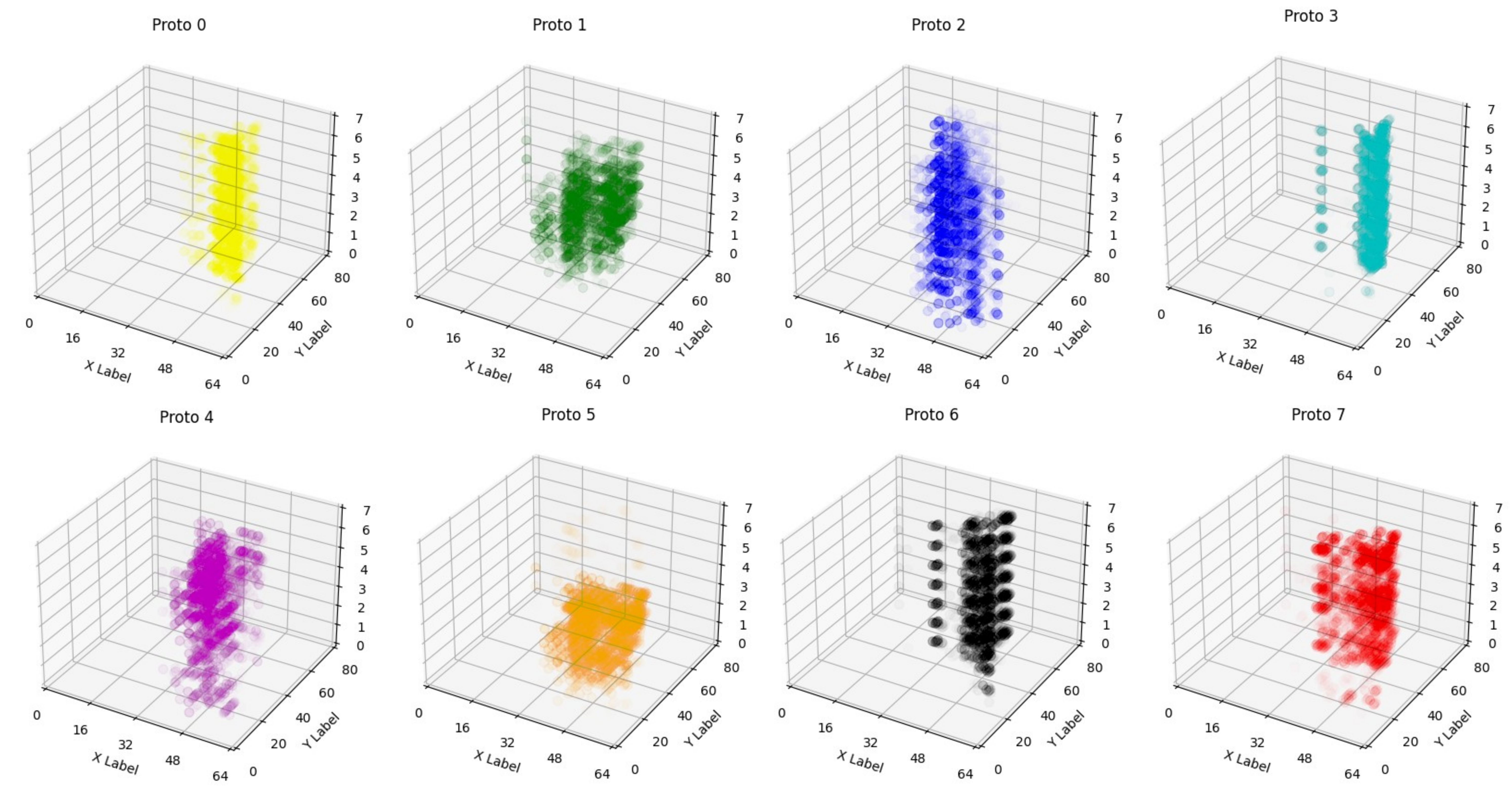}
   \caption{Chicago: Mapping Similar Samples to Every Prototype over Space and Time}
   \label{geo: chicago geopro4}
\end{minipage}
\end{figure}

\subsection{GeoPro-Net Interpretability}

In this section, We identified an accident case that happened on January 15, 2021, Chicago. GeoPro-Net is interpreted by identifying the most similar training samples to the learned prototypes, and then using the linear relationships between them and the occurrence of the event of interest. Specifically, we locate the most similar training samples to the learned prototype $P$, and the visualization of such sample will be presented as the visualization of this prototype. Figure \ref{geo: chicago geopro2} illustrates the explanations by GeoPro-Net for the cases in Chicago. In this particular case with an event, the associated concepts include high traffic volumes and crowd flows, which are similar to the concepts of positive prototypes. The higher similarity score multiplied by higher coefficients contributes to the prediction of an event. Conversely, blue prototypes with concepts, including low traffic volume and speed indicate a safe environment, contributing less to the occurrence of an event.

To understand how our learned prototypes can capture heterogeneous patterns over time and space, We map the 
top $1\%$ most similar samples to each learned prototype in the study area as illustrated in Figure \ref{geo: chicago geopro3}. We plot corresponding heat maps of Chicago for 8 learned prototypes, where the hot area indicates a higher number of top $1\%$ most similar samples. In other words, each prototype captures different patterns represented by hot regions over the heterogeneous space. For example, the learned prototype $P_3$ associated with higher accident risk are both represented in the red box of Figure \ref{geo: chicago geopro2} and $4th$ heat map of Figure \ref{geo: chicago geopro3}. It is the most populated area in Chicago. Oppositely, the learned prototype $P_1$ associated with less accident risk is represented by the blue box and the $2nd$ heat map, which represents the less-populated surrounding area. Therefore, we can visualize the heterogeneous patterns through the distribution of learning prototypes, and the positive and negative correlations between prototypes and the occurrence of events become self-explainable. Furthermore, to understand how prototypes are learned to be close to different samples over time, we plot the top $1\%$ most similar samples over space and different weekdays as shown in Figure \ref{geo: chicago geopro4}. 
Prototype $P_6$ covers different days of the week. Differently, prototype $P_5$ covers most of the study area but focuses on Monday and Tuesday. This may be because accident risks are generally higher at the beginning of the week as people drive to work. Therefore, this not only proves that GeoPro-Net can capture heterogeneous patterns over time, but also that GeoPro-Net has strong interpretability by mapping diverse concept prototypes to the study area.

\begin{figure}
\centering
\begin{minipage}{0.45\textwidth}
   \includegraphics[width=1\linewidth]{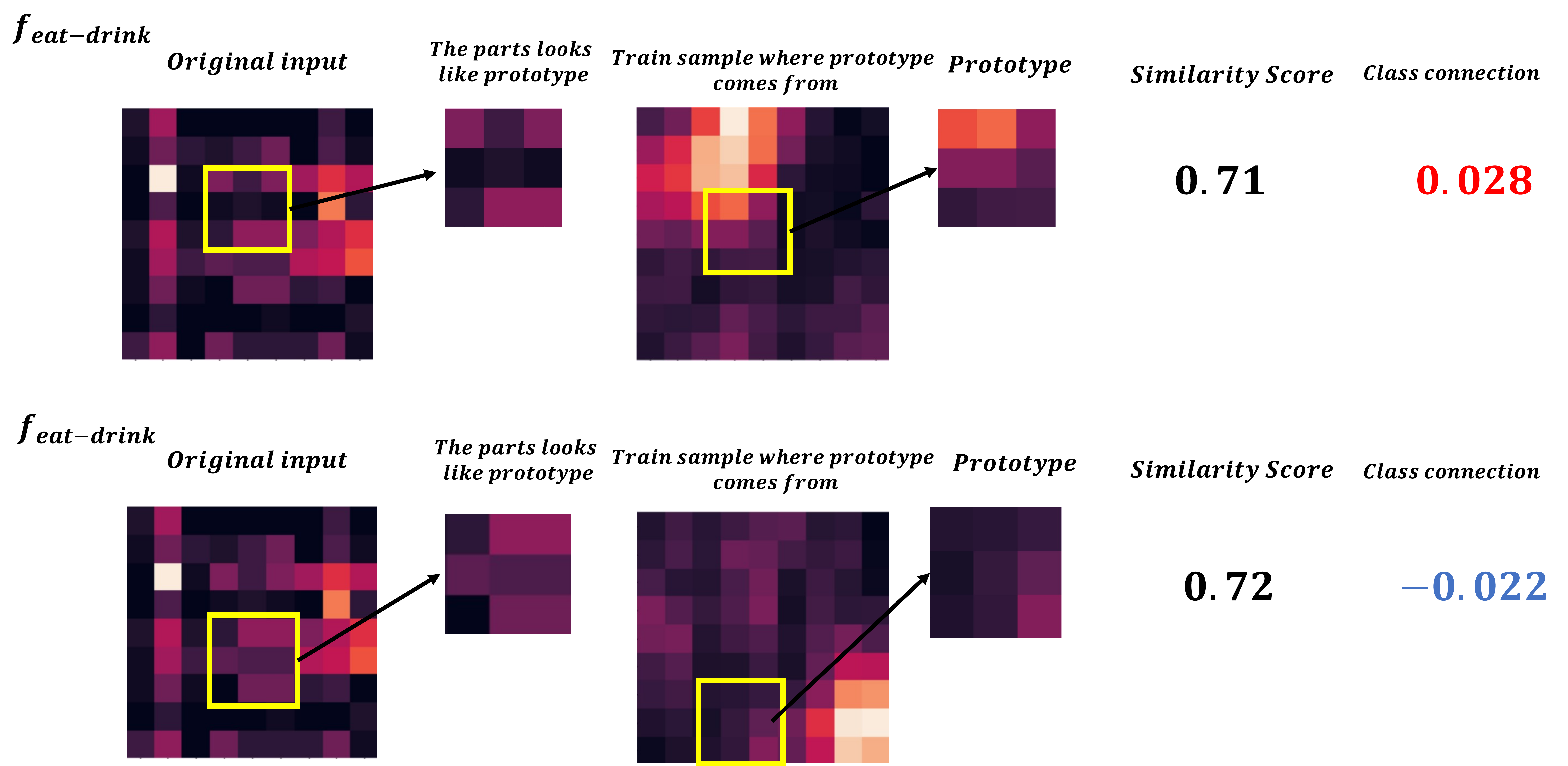}
\end{minipage}
\caption[]{Explanations by ProtoPNet, Chicago}
\label{ProtoP: chicago}
\end{figure}

\textbf{ProtoPNet} is originally designed for image classification problem, explaining its learned prototypes by visualizing a part of image with colors such wings of a bird. However, visualizing a part of study area with high-dimensional spatiotemporal features is not naturally understandable. To visualize its case study on our problem, we have to select one of the feature dimensions such as recreation facilities to plot the heat map as shown in Figure. \ref{ProtoP: chicago}. The leftmost figure is the heat map of the input region. the two small maps represent the parts of regions from input region and prototype region, where ProtoPNet thinks they are similar. The first row indicates a hot sub-region related to a hot prototype region leading to a positive linear relationship to the occurrence of events. Oppositely the recognized region in the second row is a cold area and similar to a cold prototype resulting in negative coefficients to the events. Again, the explanations of ProtoPNet only can reason the geographic correlation to the occurrence of events, but it lacks insights on causality between events and different features. Again, we found that visualizing all feature dimensions is challenging on ProtoPNet, and relying on a single feature for a heat map may lead to a loss of substantial interoperable information.

\vspace{-1mm}
\section{Conclusion}

The spatiotemporal event forecasting is essential for public safety and city management. 
While deep neural networks have superior prediction accuracy, they struggle to intrinsically interpret the complex spatial-temporal features involved. Presenting predicted scores alone is insufficient for public understanding and future urban planning. GeoPro-Net, our proposed novel solution, addressed these challenges by introducing statistical tests on spatiotemporal event data to extract meaningful concepts. It enhanced interpretability through channel fusion and geographic-based pooling, condensing concepts for clearer insights. GeoPro-Net gained better interpretability by learning prototypes of concepts and projecting them onto real-world events scenarios. 
GeoPro-Net demonstrated superior interpretability and competitive performance compared to other baselines.

\bibliography{sample-base}

\section{Appendix A}

\subsection{Feature Engineering}
In this section, we explain how we generate features. The data sources used to generate features are different in Chicago and New York City. \\

\textbf{Feature Summary:} Our framework deals with three types of input features: spatial, temporal, and spatiotemporal. In the Chicago dataset, we extract 31 features as the input, including 12 temporal features(e.g., temperature, precipitation), 13 spatial features (e.g., total road length, avg. speed limit), and 6 spatiotemporal features (e.g., traffic volume). In the New York City dataset, we extract 37 features as input, including 15 temporal features, 18 spatial features, and 4 spatial-temporal features.

\textbf{Chicago ------------------------------------------------------------}
\textbf{Temporal Features} $F_T$ Such Weather features are generated from the date of Vehicle Crash or Crime Records, where all grid cells share a vector of temporal features in a time interval. Weather features include temperature, precipitation, snowfall, wind speed, etc.  \\
\textbf{Spatial Features} $F_S$ are generated based on each grid cell and remain the same over different time intervals. First, POI features are the number of POI data in each grid cell for different categories. For example, one of the POI types is shopping, we count the number of shopping instances in each grid cell. Second, basic road condition features are extracted from road network data, in which we calculate the summation or average of provided data for road segments in each grid cell. \\
\textbf{Spatio-Temporal Features} $F_{ST}$ such as real-time traffic conditions are estimated by taxi GPS data and Bus GPS data. Spatio-temporal features include pick-up volumes, drop-off volumes, traffic speed, etc.

\noindent\textbf{Feature Summary} In total, 36 features are extracted, including 12 temporal features, 18 spatial features, and 6 spatial-temporal features for each location $s$ and time interval $t$.

\textbf{New York City -----------------------------------------------}
\textbf{Temporal Features} $F_T$ includes humidity, wind speed, temperature, precipitation, snowfall etc.  \\
\textbf{Spatial Features} $F_S$ are POI features including residential, education facility, cultural facility, recreational facility, social services, transportation facility, commercial, government facility, religious institution, health services, public safety, water, miscellaneous, total number of roads, total mileage, highway mileage, bridge count, tunnel count. \\
\textbf{Spatio-Temporal Features} $F_{ST}$ such as real-time traffic conditions are estimated by taxi GPS data including yellow taxi pick-up\&drop-off volumes, green taxi pick-up\&drop-off volumes.

\noindent\textbf{Feature Summary} In total, 37 features are extracted, including 15 temporal features, 18 spatial features, and 4 spatial-temporal features for each location $s$ and time interval $t$.

\begin{figure*}[t]
 \centering
 \includegraphics[width = 0.8\textwidth]{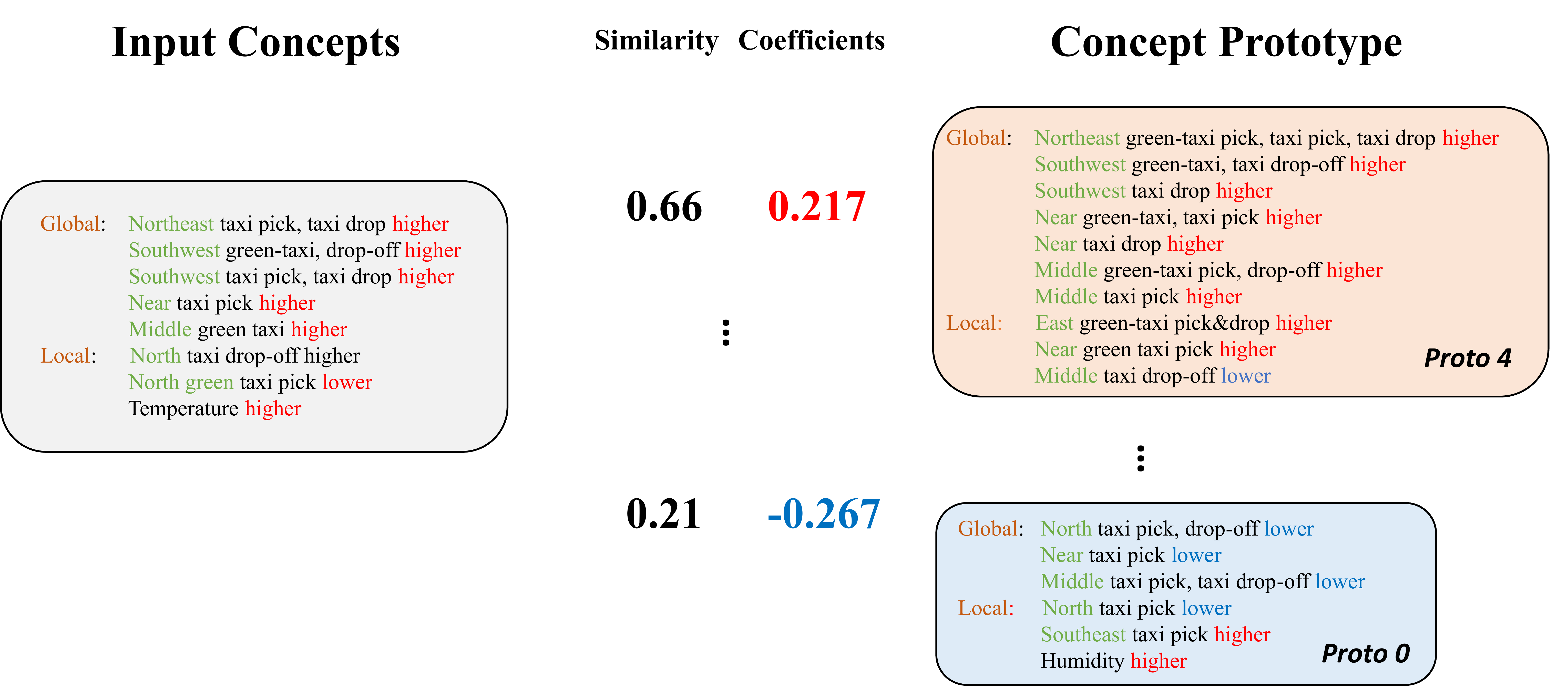}
 \caption{Grey boxes represent the encoded concepts for this given case, and red and blue boxes represent learned concept prototypes with positive and negative correlations to the occurrence of events.}
 \label{fig:Ng1}
\end{figure*}

\subsection{Symbol Table}
\setlength{\extrarowheight}{.5em}
\begin{tabular}{ |p{1.5cm}|p{6cm}|  }
\hline
\multicolumn{2}{|c|}{Symbol Table} \\
\hline
Symbol & Explainations \\

\hline
$\S$ & Spatial filed, study area \\
$s$  &  A partitioned location, grid cell   \\
$T$  &    Temporal filed, study period \\
$t$  &    Time interval (e.g. hours, days) \\
$m$  &    Width of study area \\
$n$  &    Length of study area \\
$F_T$  & Temporal features (weather, time)    \\
$F_S$  &   Spatial features (e.g. POI) \\
$F_{ST}$  &    Spatiotemporal features (e.g. traffic conditions)\\
$Y$  &    Event (binary) \\ 
$\Psi_L$  &   Local Significance Test\\ 
$\Psi_G$  &   Global Significance Test\\ 
$q$  &  Number of pooling regions   \\
$\Lambda$  &  Scanning window   \\
$\omega$  &  concept convolution window sizes  \\
$C$  &  Geo-concept encoded sample  \\
$P$  &  Prototypes  \\
$K$  &  Number of prototypes  \\

\hline
\end{tabular}

\section{Appendix B}

\setlength{\extrarowheight}{0.em}
\begin{table}[t]
\begin{center}
\begin{threeparttable}[b]
\caption{Ablation Study}
\label{tab:ablation}
\begin{small}
\begin{sc}
\begin{tabular}{p{2.0cm}p{0.9cm}p{0.9cm}p{0.9cm}p{0.9cm}p{0.9cm}}
\toprule

\multirow{1}{*}{\thead{\textbf{New York City}}} &
\multicolumn{5}{c}{\thead{Crime}} \\
\cmidrule(lr){2-6}

& \footnotesize{CrsEnt} & \footnotesize{ACC} & \tiny{Precision} & \footnotesize{Recall} & \footnotesize{F1} \\
\midrule
No Pooling & .56$\pm$1.2\tiny\textperthousand  & .77$\pm$0.9\tiny\textperthousand & .63$\pm$1.7\tiny\textperthousand & .24$\pm$0.4\tiny\textperthousand & .37$\pm$0.8\tiny\textperthousand  \\
Max Pooling & .52$\pm$0.7\tiny\textperthousand  & .77$\pm$0.7\tiny\textperthousand & .67$\pm$0.6\tiny\textperthousand & .26$\pm$0.6\tiny\textperthousand & .41$\pm$1.0\tiny\textperthousand  \\
Spatial* & \textbf{.51}$\pm$0.3\tiny\textperthousand  & \textbf{.79}$\pm$0.1\tiny\textperthousand & \textbf{.68}$\pm$0.4\tiny\textperthousand & \textbf{.27}$\pm$0.4\tiny\textperthousand & \textbf{.43}$\pm$0.2\tiny\textperthousand  \\

\multirow{1}{*}{\thead{ }} &
\multicolumn{5}{c}{\thead{Accident}} \\
\cmidrule(lr){2-6}

& \footnotesize{CrsEnt} & \footnotesize{ACC} & \tiny{Precision} & \footnotesize{Recall} & \footnotesize{F1} \\
\midrule
No Pooling & .59$\pm$2.8\tiny\textperthousand  & .73$\pm$3.1\tiny\textperthousand & .45$\pm$1.1\tiny\textperthousand & .23$\pm$0.5\tiny\textperthousand & .30$\pm$0.9\tiny\textperthousand \\
Max Pooling & .56$\pm$0.6\tiny\textperthousand  & .75$\pm$1.2\tiny\textperthousand & .48$\pm$0.8\tiny\textperthousand & \textbf{.28}$\pm$0.3\tiny\textperthousand & .36$\pm$0.5\tiny\textperthousand \\
Spatial* & \textbf{.54}$\pm$0.4\tiny\textperthousand  & \textbf{.76}$\pm$0.2\tiny\textperthousand & \textbf{.50}$\pm$0.6\tiny\textperthousand & \textbf{.28}$\pm$0.2\tiny\textperthousand & \textbf{.37}$\pm$0.4\tiny\textperthousand  \\
\bottomrule
\end{tabular}
\begin{tablenotes}
    \item *: Spaital Concept Pooling
    \item $'$: $\times10^{-3}$
\end{tablenotes}
\end{sc}
\end{small}
\end{threeparttable}
\end{center}

\label{table: ablation2}
\end{table}

\subsection{Ablation Study}

We examine the effects of removing spatial pooling or replacing with max pooling in the GeoPro-Net. Table. 4 shows that GeoPro-Net achieves optimal performance with spatial concept pooling. Replacing spatial pooling by max pooing slightly lowers the performance and loses the capability to interpret based on geographic information. Removing the pooling layer results in a dramatic decrease in its performance due to noisy information.

\begin{figure}
\centering
\begin{minipage}[c]{0.45\textwidth}
\centering\includegraphics[width=1\textwidth]{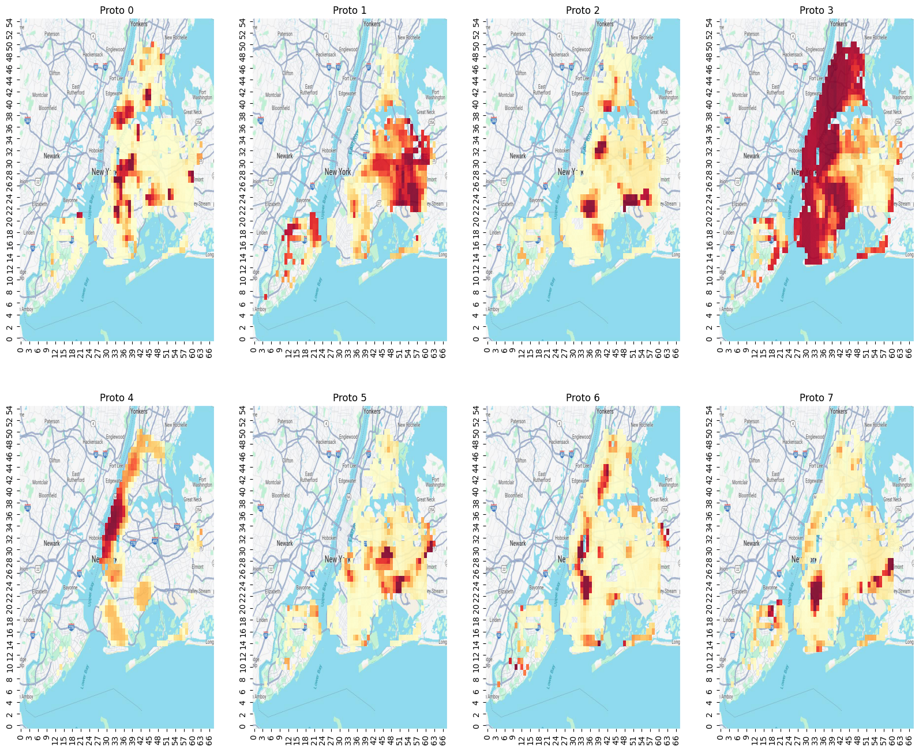}
   \caption{NYC: Mapping Similar Samples to Every Prototype over Space}
    \label{fig:Ng2}
\end{minipage}
\begin{minipage}[c]{0.45\textwidth}
\centering\includegraphics[width=1\textwidth]{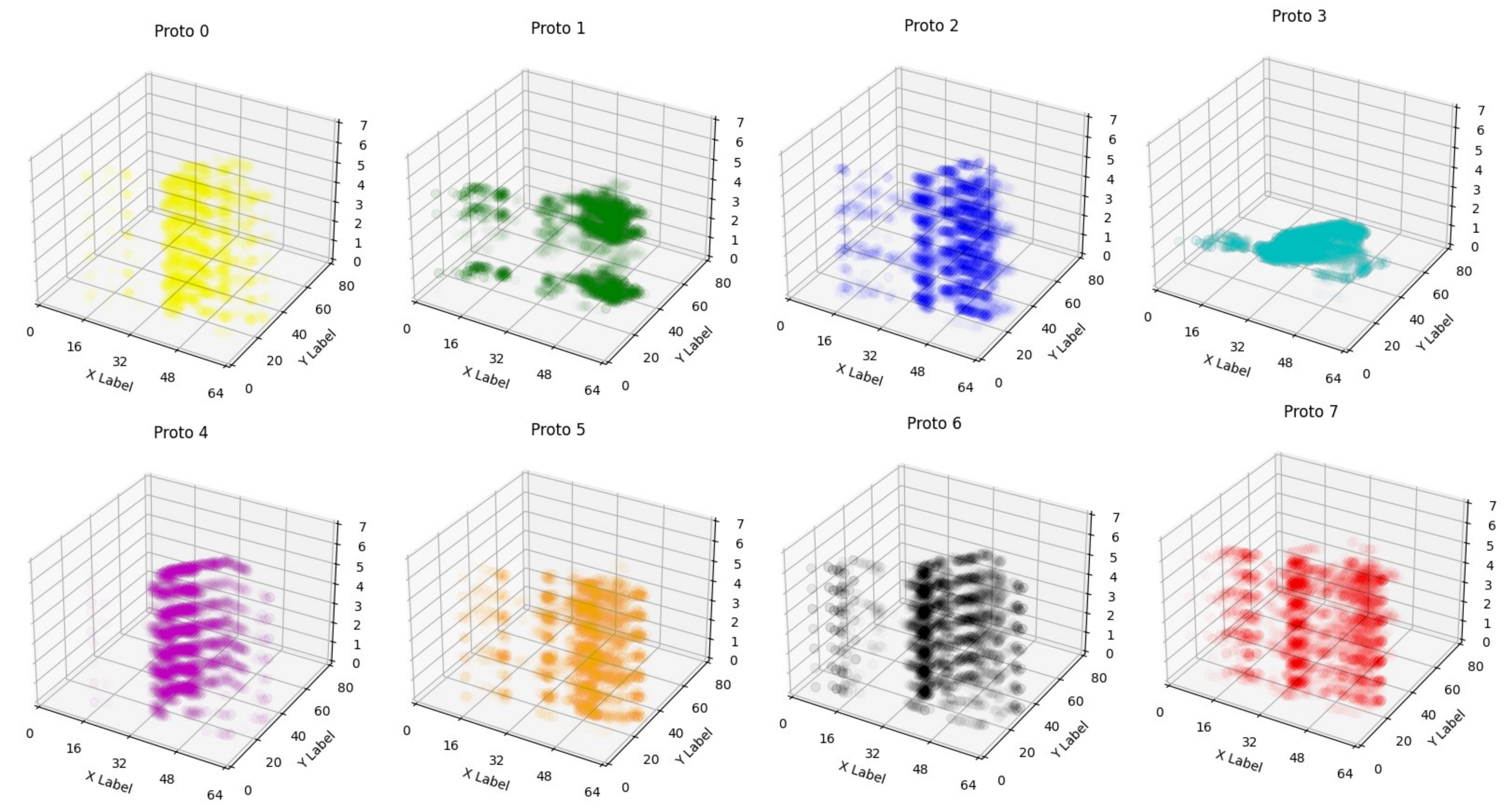}
   \caption{NYC: Mapping Similar Samples to Every Prototype over Space and Time}
   \label{fig:Ng3}
\end{minipage}
\end{figure}

\subsection{GeoPro-Net} 

We exemplify an extra case and identify a crime case that happened on Jan 9th, 2021, in New York City. All compared methods are studied in the same cases. We can observe a similar pattern in Figure. \ref{fig:Ng2}. This case in Chicago with an event tends to be associated with concepts such as high taxi pick-ups and drop-offs, which are similar to the concepts of prototypes and contribute to the prediction of an event. Similar to the case in Chicago, blue prototypes with concepts including low taxi pick-ups and drop-offs indicate contribute less to the occurrence of an event. 

To understand how our learned prototypes can capture heterogeneous patterns over time and space. We map the 
top $1\%$ most similar samples to learned prototypes in the study area as illustrated in Figure. \ref{fig:Ng2}. We plot corresponding heat maps of New York City for 8 learned prototypes, where the hot area indicates a higher number of top $1\%$ most similar samples. In other words, each prototype captures different patterns represented by hot regions over the heterogeneous space. For example, the learned prototype $P_4$ associated with higher crime risk is both represented in the red box of Figure. \ref{fig:Ng2} and $5th$ heat map of Figure. \ref{fig:Ng2}. We found that is the populated area of Manhattan. Oppositely, Learned prototype $P_4$ associated with less crime risk is represented by the blue box and the $1th$ heat map, where most regions are less-populated surrounding areas. Furthermore, to understand how prototypes are learned to be close to different samples over time, we plot the top $1\%$ most similar samples over space and different weekdays as shown in Figure. \ref{fig:Ng3}. We can observe that prototypes are not learned to be evenly distributed over time but have preferences. Prototype 3 covers most of the study area but focuses on Monday.

\subsection{ProtoPNet} 
Visualizing a portion of the study area with high-dimensional spatiotemporal features cannot be naturally understood by the user. To visualize its case study on our problem, we have to select one of the feature dimensions such as recreation facilities to plot the heat map as shown in Figure. \ref{ProtoP: NYC}. The leftmost figure is the heat map of the input region. the two small maps represent the parts of regions from input region and prototype region, where ProtoPNet thinks they are similar. The first row indicates a hot sub-region related to a hot prototype region leading to a positive linear relationship to the occurrence of events. Oppositely the recognized region in the second row is a cold area and similar to a cold prototype resulting in negative coefficients to the events. Again, the explanations of ProtoPNet only can reason the geographic correlation to the occurrence of events, but it lacks insights on causality between events and different features. Again, we found that visualizing all feature dimensions is challenging on ProtoPNet, and relying on a single feature for a heat map may lead to a loss of substantial interoperable information. 

\begin{figure}
\centering
\begin{minipage}{0.45\textwidth}
   \includegraphics[width=1\linewidth]{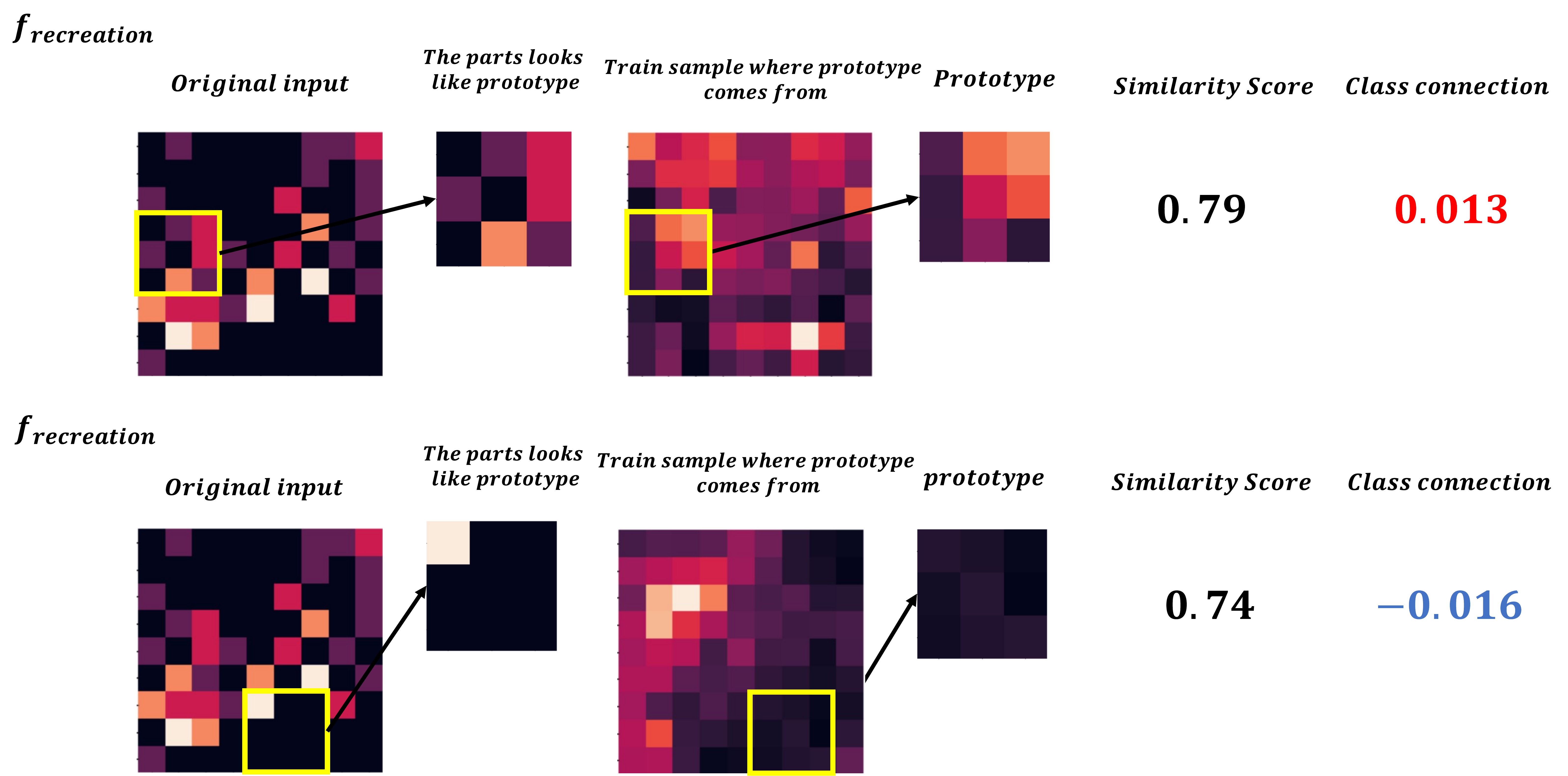}
\end{minipage}
\caption[]{Explanations by ProtoPNet, New York City}
\label{ProtoP: NYC}
\end{figure}

\begin{figure}
\centering
\begin{minipage}{0.12\textwidth}
   \includegraphics[width=1.0\linewidth]{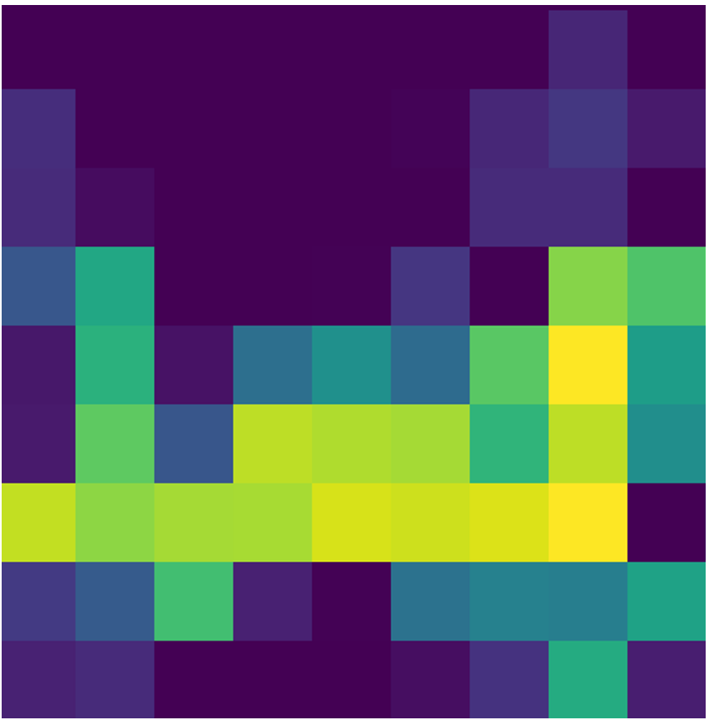}
\end{minipage}
\hspace{1.5cm} 
\begin{minipage}{0.12\textwidth}
   \includegraphics[width=1.0\linewidth]{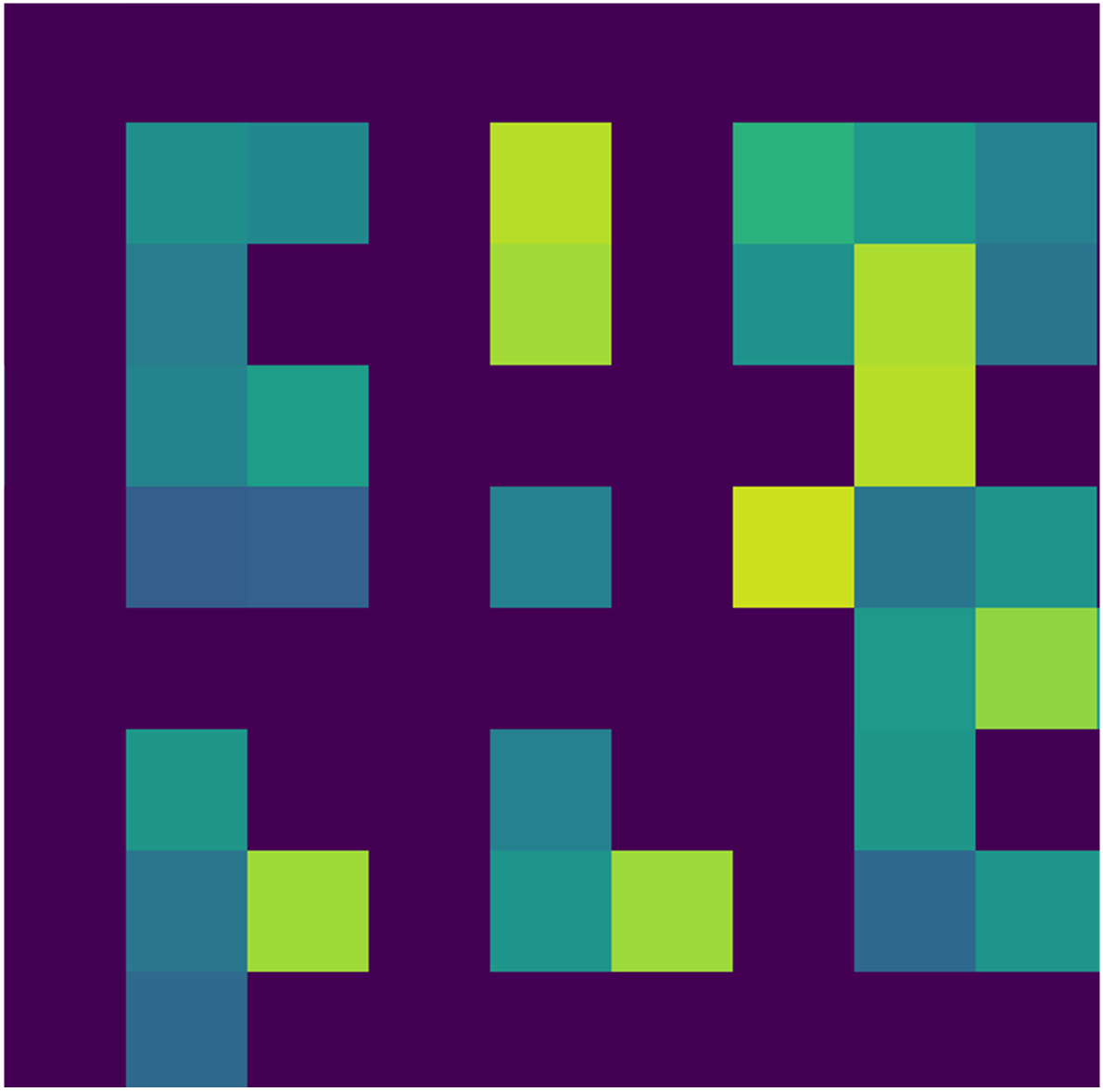}
\end{minipage}
\label{ASTGCN}
\caption[]{Attention Map by ASTGCN in NYC (Left) and Chicago (Right)}
\end{figure}

\subsection{ASTGCN} is a partially interpretable method relying on an attention mechanism, where a generated attention map can visualize how much attention the model pays to different locations. In Fig. 12, ASTGCN pays more attention on those lighted regions. However, ASTGCN neither tells users the relationships of those focused regions to the occurrence of events nor explains how different factors lead to the happening of events. Therefore, ASTGCN barely brings meaningful insights and practical benefits to real-world city management.

\end{document}